\documentclass{article} 
\usepackage{iclr2025_conference,times}

\usepackage{amsmath,amsfonts,bm}

\def\Figref#1{Figure~\ref{#1}}

\def\eqref#1{equation~\ref{#1}}
\def\Eqref#1{Equation~\ref{#1}}

\def\1{\bm{1}}

\DeclareMathAlphabet{\mathsfit}{\encodingdefault}{\sfdefault}{m}{sl}
\SetMathAlphabet{\mathsfit}{bold}{\encodingdefault}{\sfdefault}{bx}{n}

\usepackage{hyperref}
\usepackage{url}

\usepackage{amsmath} 
\usepackage{subfigure}
\usepackage{graphicx}
\usepackage{wrapfig} 
\title{FlowDreamer: Exploring High Fidelity Text-to-3D Generation Via Rectified Flow}

\author{Hangyu Li \thanks{Equal contribution. } \thanks{Work done during intern at the Alibaba Group}  \\
The Hong Kong University of Science
and Technology (Guangzhou) \\
\texttt{cyjdlhy@gmail.com} 
\And
Xiangxiang Chu $^*$ \\
Alibaba Group \\
\texttt{chuxiangxiang.cxx@alibaba-inc.com}
\And
Dingyuan Shi  \\
Alibaba Group\\
\texttt{dingyuan.shi@outlook.com}
\And
Lin Wang \thanks{Corresponding author} \\
The Hong Kong University of Science
and Technology (Guangzhou) \\
The Hong Kong University of Science
and Technology \\
\texttt{linwang@hkust.hk, alwang.hkust@gmail.com}
}

\newcommand{\eg}{\textit{e}.\textit{g}.}

\newcommand{\ie}{\textit{i}.\textit{e}.}
\newcommand{\etc}{\textit{etc}}

\ifodd 1

\else

\fi
\newcommand{\fakeparagraph}[1]{\vspace{1mm}\noindent\textbf{#1.}}

\iclrfinalcopy 
\begin{document}
\maketitle

\begin{abstract}
Recent advances in text-to-3D generation have made significant progress. In particular, with the pretrained diffusion models, existing methods predominantly use Score Distillation Sampling (SDS) to train 3D models such as Neural Radiance Fields (NeRF) and 3D Gaussian Splatting (3D GS).
However, a hurdle is that they often encounter difficulties with over-smoothing textures and over-saturating colors.
The rectified flow model -- which utilizes a simple ordinary differential equation (ODE) to represent a straight trajectory -- shows promise as an alternative prior to text-to-3D generation. 
It learns a \textit{time-independent} vector field, thereby reducing the ambiguity in 3D model update gradients that are calculated using \textit{time-dependent} scores in the SDS framework.
In light of this, we first develop a mathematical analysis to seamlessly integrate SDS with rectified flow model, paving the way for our initial framework known as Vector Field Distillation Sampling (VFDS).
However, empirical findings indicate that VFDS still results in over-smoothing outcomes.
Therefore, we analyze the grounding reasons for such a failure from the perspective of ODE trajectories.
On top, we propose a novel framework, named \textbf{FlowDreamer}, which yields high-fidelity results with richer textual details and faster convergence.
The key insight is to leverage the coupling and reversible properties of the rectified flow model to search for the corresponding noise, rather than using randomly sampled noise as in VFDS.
Accordingly, we introduce a novel Unique Couple Matching (\textbf{UCM}) loss, which guides the 3D model to optimize along the same trajectory.
Our FlowDreamer is superior in its flexibility to be applied to both NeRF and 3D GS.
Extensive experiments demonstrate the high-fidelity outcomes and accelerated convergence of FlowDreamer. 
Moreover, we highlight the intriguing open questions, such as initialization challenges in NeRF and sampling techniques, to benefit the research community.  
For more details, please visit our project page at \url{https://vlislab22.github.io/FlowDreamer}.
\end{abstract}

\begin{figure}[t!]
    \centering
    \includegraphics[width=1.\linewidth]{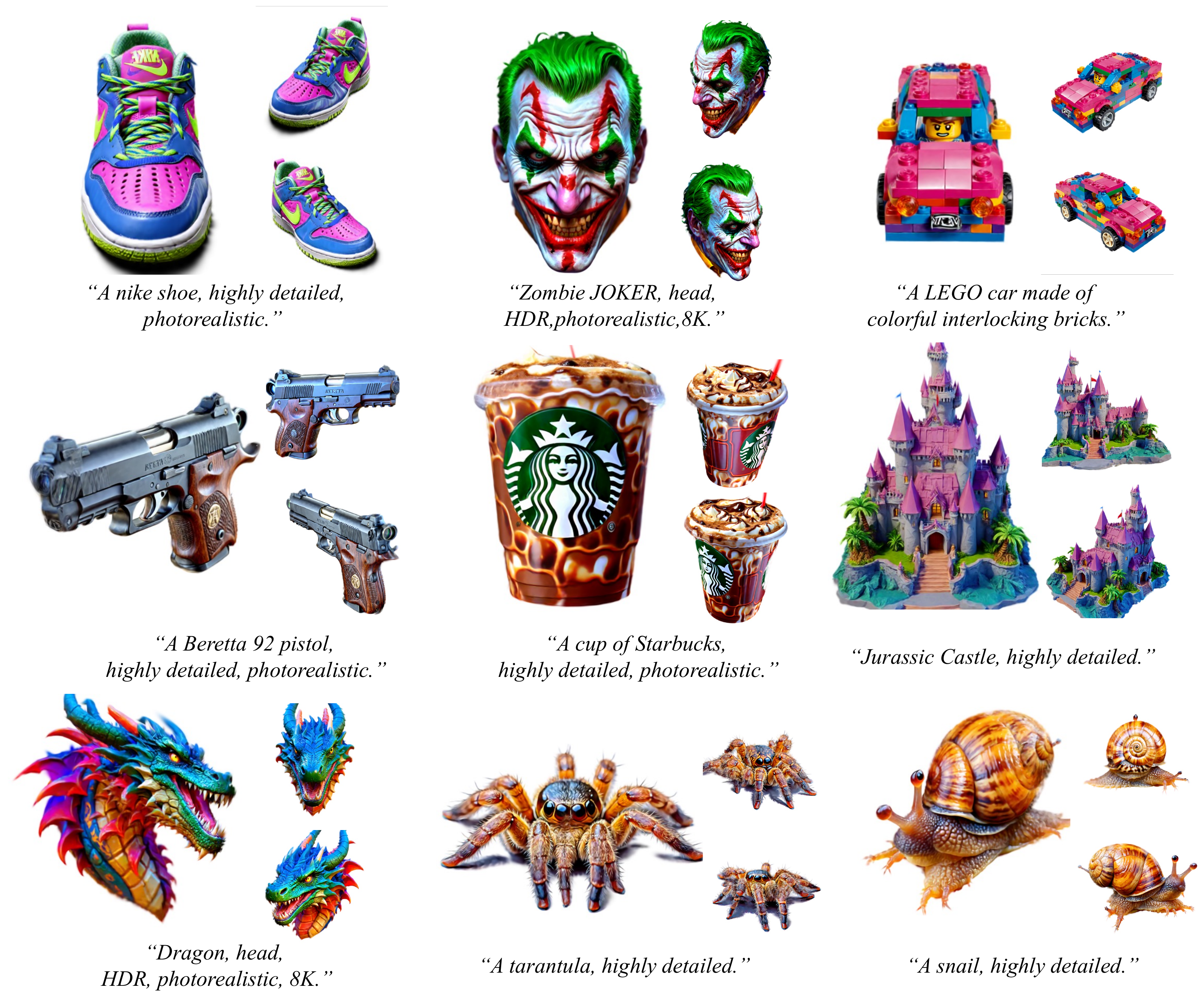} 
     \vspace{-16pt}
    \caption{FlowDreamer uses a pretrained rectified flow model to generate high-fidelity results from text prompts. It can generate not only highly realistic objects, such as guns and shoes, but also fantastical ones, such as dragon heads.}
    \label{fig:cover}
    \vspace{-10pt}
\end{figure}


\vspace{-15pt}
\section{Introduction}
3D generation enjoys broad applications in diverse fields, such as the Metaverse, games, education, architecture design, and films, and has attracted significant research interest recently~\citep{xie2024latte3d, wang2024crm, tang2024mvdiffusion++, dreamfusion, chen2023fantasia3d, lin2023magic3d, jain2022dreamfield, tang2023dreamgaussian, jiang2024general}. Text-to-3D generation -- which generates 3D contents from user-input text -- has emerged as one of the promising 3D generation paradigms due to its ease of use~\citep{wang2023sjc, yi2023gaussiandreamer, wang2024prolificdreamer, nichol2022point_e, jun2023shap_e}. 

Recently, with the advances of text-to-2D image synthesis techniques based on the diffusion models, text-to-3D generation also undergoes a surge of research interest.
A seminal work, DreamFusion~\citep{dreamfusion} sets a cornerstone by proposing Score Distillation Sampling (SDS) to address this issue by leveraging pretrained text-to-image diffusion model, to train Neural Radiance Fields (NeRF)~\citep{nerf}.
It has been rapidly evolved to 3D Gaussian Splatting (3D GS)~\citep{3dgs,tang2023dreamgaussian,yi2023gaussiandreamer} for faster training and rendering.

 
Despite the success, existing works ~\citep{lin2023magic3d, dreamfusion, zhu2023hifa} unveil that SDS suffers from issues such as over-smoothing textures and over-saturating colors.
For this reason, some attempts, \eg,\cite{wang2024prolificdreamer,liang2023luciddreamer,wu2024consistent3d} improve SDS from different perspectives. 
For example,
ProlificDreamer~\citep{wang2024prolificdreamer} introduces variational score distillation (VSD), which models 3D parameters as random variables to distill 3D assets.
However, it requires significantly more time to optimize. %
Consistent3D~\citep{wu2024consistent3d} designs a consistency distillation sampling method to train the 3D model. Nevertheless, the quality improvements are still limited.
LucidDreamer~\citep{liang2023luciddreamer} proposes interval score matching (ISM) loss in the diffusing trajectory, but the loss is formulated based on strong assumptions and  drops many terms with the same scale.


Recently, flow matching approaches~\citep{liu2022rcflow,lipman2022flowmatching} pave new ways for fast and high-quality generation.
Among them, rectified flow model~\citep{liu2022rcflow,esser2024sd3,liu2023instaflow} uses a simple ordinary differential equation (ODE) to represent a straight trajectory. 
It trains a \textit{time-independent} vector field, indicating that the direction of the vector field remains constant. 
Whereas diffusion trajectories result in curved trajectories~\citep{lipman2022flowmatching}.
Score~\citep{song2020score} is \textit{time-dependent}, meaning that SDS optimizing over uniformly sampled values of $t$ can produce different gradient directions. Owning to its merits,  we interestingly find that it 
could serve as an alternative prior for text-to-3D generation. 

\begin{wrapfigure}{r}{7cm}
\centering
\vspace{-23pt}
\includegraphics[width=0.72\linewidth]{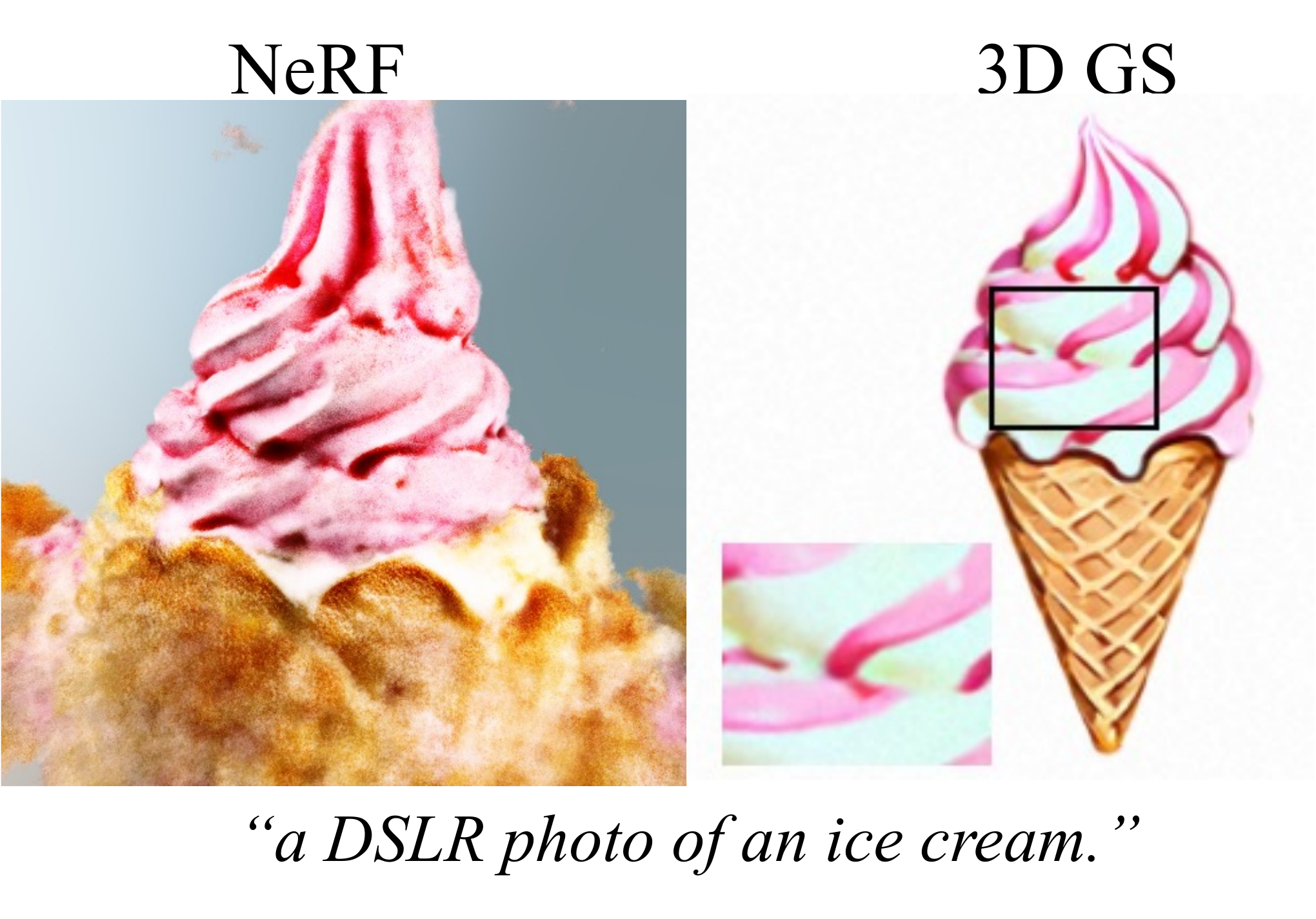} 
\vspace{-18pt}
\caption{An example of over-smoothing results.}
\label{fig:over_soothing}
\end{wrapfigure}

In light of this, we first develop a mathematical analysis to seamlessly integrate SDS with rectified flow model. 
This enables us to build up an initial framework, named as Vector Field Distillation Sampling (VFDS). 
However, empirical results demonstrate that VFDS still leads to over-smoothing textures (See~\Figref{fig:over_soothing}).
To this end, we further analyze the grounding reasons for such a failure from the perspective of ODE trajectories.
This way, we find that, \textit{as VFDS randomly samples noise, it leads to multiple ODE trajectories in 
nearly the same images}, \ie, from camera poses with mild differences (See~\Figref{fig:trajectory}).
This causes inconsistent update directions, leading to over-smoothing textural issues.

Buttressed by the analysis, we propose a novel framework, named \textbf{FlowDreamer}, which yields high-fidelity results with rich textual details and faster convergence.
\textit{The key idea of FlowDreamer is to leverage the coupling and reversible properties of rectified flow model. }
Importantly, the reversible property is explored to search the corresponding noise from the image while the coupling property ensures that the corresponding noise is unique.
For formality, we define it as a \textit{push-backward} process to avoid the aforementioned over-smoothing issue caused by multiple ODE trajectories and better make our update directions consistent.
Empirical experiments show that the \textit{push-backward} process is efficient as three Euler discretization steps are sufficient for it to achieve plausible performance, see~\Figref{fig:cover}.%
 Accordingly, we propose a novel Unique Couple Matching (\textbf{UCM}) loss that guides the 3D model to learn the same trajectory.
Our FlowDreamer also enjoys high flexibility as it can be applied to either NeRF or 3D GS generation settings. 

We conduct extensive experiments under diverse generation settings, demonstrating high-fidelity results with rich details and faster convergence of FlowDreamer, as shown in~\Figref{fig:cover}.  
When exploring the application of FlowDreamer to NeRF, we observe some interesting phenomena. 
Moreover, we identify some intriguing open questions. First is the initialization problem that emerges when applying FlowDreamer to NeRF. This is because the distribution of the initialized image from NeRF is undefined in the Rectified flow diffusion space. We use a warm-up strategy to mitigate this issue. Secondly, the \textit{push-backward} process with different numbers of function evaluations (NFE) and sampling methods can generate some interesting results. 

In summary, our major contributions are as follows:
\begin{itemize}
\item 

We are the \textit{first} to explore a new direction by leveraging the rectified flow model as an alternative prior for text-to-3D generation. We accordingly build a mathematical analysis to adapt SDS to rectified flow model, paving the way for an initial framework -- VFDS.
Empirical results demonstrate that VFDS still leads to over-smoothing. We further analyze the underlying reasons for this issue from the perspective of ODE trajectories.
\item 
Based on the analysis, we further propose a text-to-3D framework, FlowDreamer, with a novel UCM loss. The loss 
is build opon the \textit{push-backward} process to search for corresponding noise, rather than using randomly sampled noise in VFDS.
\item
Extensive experiments in both NeRF and 3D GS generation settings demonstrate high-fidelity results with rich details and faster convergence for our FlowDreamer. We also identify some interesting open questions, such as initialization issues for NeRF and sampling techniques in the noise search process.


\end{itemize}

\section{Related Works}
\fakeparagraph{Text-to-3D generation}
It aims to create 3D assets from user-input text.
DreamFusion~\citep{dreamfusion} proposes Score Distillation Sampling (SDS) that leverages the pretrained diffusion models to train a NeRF.
However, SDS suffers from issues such as over-smoothing textures, low resolution, slow convergence, multi-faced problem~\citep{wang2024prolificdreamer,lin2023magic3d,dreamfusion}, \etc.
Magic3D~\citep{lin2023magic3d} designs a coarse-to-fine two-stage training pipeline and changes the 3D model to DMtet~\citep{shen2021dmtet} to improve the resolution of the generated 3D results.
Later on, some works~\citep{tang2023dreamgaussian, yi2023gaussiandreamer, liang2023luciddreamer,chen2023gaussiansplatting,jiang2024general,li2024dreamscene,jiang2024brightdreamer} take 3D GS as the 3D model for faster training and rendering.
Recently, to overcome the multi-face problem, some works~\citep{shi2023mvdream, wang2023imagedream, tang2024mvdiffusion++} fine-tune the pretrained diffusion models to generate multi-view images.

\fakeparagraph{Design variant of SDS loss}
To overcome the issue of over-smoothing textural issues, some attempts~\citep{wang2023sjc, wang2024prolificdreamer, liang2023luciddreamer, wu2024consistent3d, zhu2023hifa, katzir2023nfsd, yu2023csd} focus on designing different SDS losses. For example,
\cite{wang2023sjc} proposes Score Jacobian Chaining, which applies the chain rule to the estimated score to enhance generation quality.
ProlificDreamer~\citep{wang2024prolificdreamer} proposes VSD to model 3D parameters as random variables to distill 3D assets. 
Although with improved quality, it requires a much longer time to optimize.
Consistent3D~\citep{wu2024consistent3d} designs a consistency distillation sampling method to train the 3D model. Nevertheless, the quality improvements are limited.
LucidDreamer~\citep{liang2023luciddreamer} proposes ISM loss in the diffusing trajectory, but the loss drops many terms with the same scale.
These methods build loss towards either DDPM~\citep{ddpm} or DDIM~\citep{songddim} models. \textit{By contrast, we propose a novel UCM loss that is build opon the push-backward process to search for corresponding noise, rather than using randomly sampled noise with rectified flow-based models. Our UCM demonstrates high-fidelity results with rich details and faster convergence under either NeRF or GS generation settings. }


\fakeparagraph{Diffusion and Flow based models} 
Recent advances in text-to-image generation have witnessed a significant progress.
Diffusion models define a process, which progressively converts a distribution of training data to pure Gaussian noise. By learning the reverse process, one can sample data following the distribution.
DDPM~\citep{ddpm} employs a Markov chain to achieve the above process, while DDIM~\citep{songddim} proposes to reduce the iteration step by maintaining the marginal distribution hence free from the restriction of Markov property.
The diffusion process can be essentially modeled as stochastic differential equations (SDE)~\citep{song2020score} or ordinary stochastic equations (ODE)~\citep{lipman2022flowmatching}.
On the other hand, flow-base models~\citep{liu2022rcflow,lipman2022flowmatching,liu2023instaflow} pave new ways for faster and higher-quality generation.

Recently, rectified flow~\citep{liu2022rcflow} -- one of the ODE methods -- defines a simple process, optimizing the trajectories in diffusion space to be as straight as possible. \textit{We are the first to explore a new direction by leveraging the rectified flow model as an
alternative prior for text-to-3D generation. We propose a novel UCM loss, built upon the push-backward process to search for corresponding noise, Extensive experiments in both NeRF and 3D GS generation settings demonstrate higher-fidelity results with richer details and faster convergence}.

\begin{figure}[t!]
\centering
\includegraphics[width=1.0\linewidth]{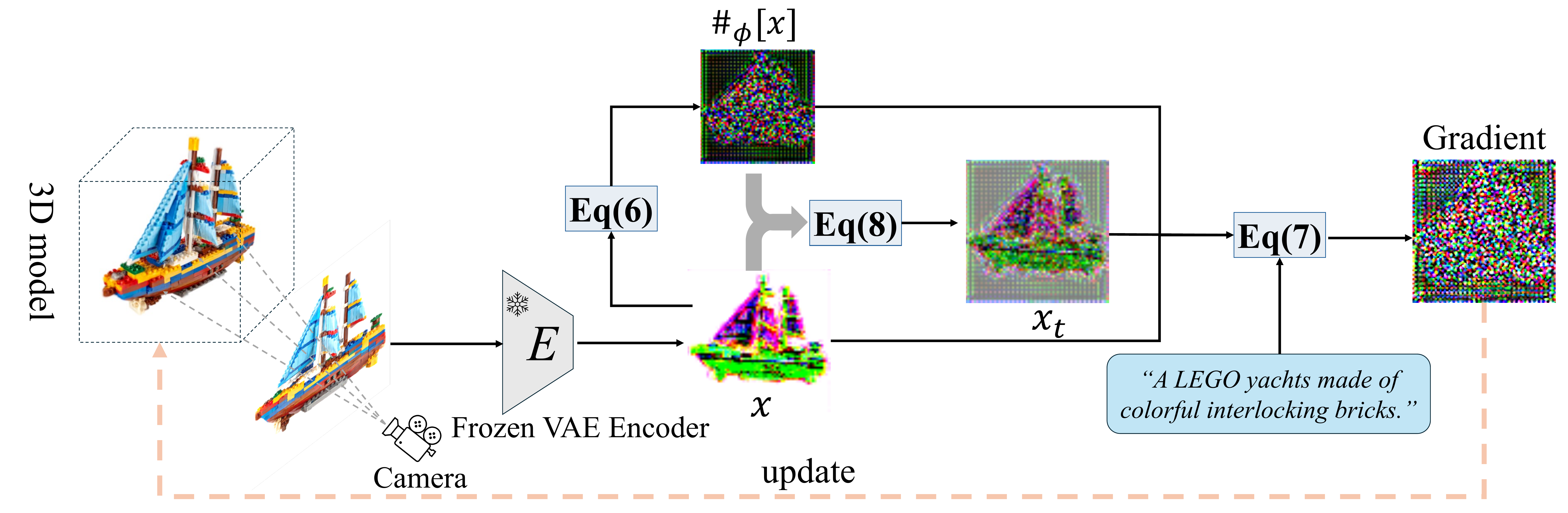}
\vspace{-20pt}
\caption{\textbf{Illustration of our FlowDreamer}. Images of random views from different camera poses are sampled and then input to the VAE encoder to obtain the latents. We replace the randomly sampled noise $\epsilon$ in VFDS with $\#_\phi[x]$ via the \textit{push-backward} process. Next, we sample $t$ from $U[0, 1]$ and interpolate to obtain $x_t$. Finally, the UCM loss with the conditional prompt is applied to update the 3D model.} 
\label{fig:pipeline}
\vspace{-10pt}
\end{figure}

\section{FlowDreamer}
\noindent \textbf{Overview.} An overview of our proposed FlowDreamer is depicted in Figure~\ref{fig:pipeline}.  The key insight is to leverage the coupling and reversible properties of the rectified flow model to search for the corresponding noise, rather than using randomly sampled noise as in our initial framework VFDS (see Sec.~\ref{sec:vfds-sec}).
Accordingly, we introduce a novel Unique Couple Matching (\textbf{UCM}) loss in Sec.~\ref{sec:ucm_loss}, which guides the 3D model to optimize along the same trajectory. Finally, our FlowDreamer can be applied to two types of 3D models: 3D Gaussian splatting~\citep{3dgs} and NeRF~\citep{nerf} settings. Now let's describe the details (see Sec.~\ref{sec:flowdreamer}).


\vspace{-6pt}
\subsection{VFDS: SDS in the Lens of Rectified flow} 
\label{sec:vfds-sec}
\fakeparagraph{Adapting SDS to the rectified flow framework}
We first briefly introduce the rectified flow~\citep{liu2022rcflow}.
Let $\pi_1$ and $\pi_0$ denote Gaussian distribution $\mathcal{N}(\mathbf{0}, \mathbf{I})$ and data distribution, respectively.
$\epsilon$ and $x_0$ are respectively sampled from $\pi_1$ and $\pi_0$.
Rectified flow defines the forward process as(to simplify the representation, below, $x_0$ and $x_t$ indicate the latent space.):
\begin{equation}
x_t = t \epsilon + (1-t) x_0,        t \in [0, 1]
\label{eq:interpolation}
\end{equation}

\begin{figure}[t!]
\centering
\includegraphics[width=1.0\linewidth]{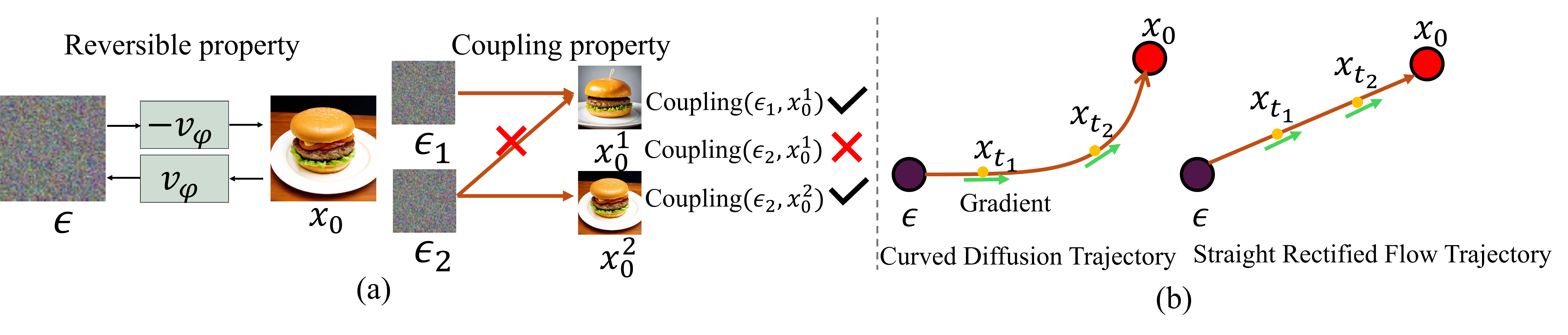}
\vspace{-18pt}
\caption{(\textbf{a}): \textbf{An illustration of the reversible and coupling properties of the rectified flow model}. The reversible property indicates that $\epsilon$ can map to $x_0$, and $x_0$ can map to $\epsilon$ by reversing the direction of $v_\phi$. The coupling property indicates that $\epsilon$ and $x_0$ can only form a unique coupling. For example, $\epsilon_2$ and $x_0^2$ form a coupling $(\epsilon_2, x_0^2)$; therefore, $\epsilon_2$ and $x_0^1$ can't form a coupling $(\epsilon_2, x_0^1)$ again. (\textbf{b}): \textbf{An illustration of the trajectories of diffusion and rectified flow.} The gradient direction of the diffusion trajectory varies with different $t$, while the rectified flow roughly remains the same for different $t$ under ideal circumstances.}
\label{fig:flow_property}
\end{figure}

Accordingly, the reverse process follows the Ordinary Differential Equation (ODE) to map $\epsilon$ to $x_0$.
\begin{equation}
d x_t = v_\phi(x_t,t) dt, t \in [0, 1],  
\end{equation}
where the velocity $v_\phi$ is estimated by a learnable network $\phi$.
The model is trained as follows:
\begin{equation}
\mathcal{L}_{\mathrm{rflow}}(\phi, \mathbf{x}) = \mathbb{E}_{x_0 \sim p_0, \epsilon \sim \mathcal{N}(\mathbf{0}, \mathbf{I}), t \sim U[0,1]} \left[ w(t) \| (\epsilon - x_0) - v_\phi(x_t, t) \|_2^2 \right], 
\label{eq:traning loss}
\end{equation} where $w(t)$ is a time-dependent weighting function, $U[0,1]$ denotes the uniform distribution within $[0,1]$. 
Because the rectified flow model is an ODE model,  
it has reversible and coupling properties.
The $\epsilon$ from the Gaussian noise distribution is uniquely coupled with the $x$ from the data distribution. 
Moreover, the rectified flow is reversible, as shown in~\Figref{fig:flow_property}(a).
Specifically,

\textbf{1) Reversible property}:
The $\epsilon$ from the Gaussian noise distribution can map to $x$, while $x$ from the data distribution can also map to $\epsilon$.

\textbf{2) Coupling property}: The $\epsilon$ is determined, and the $x$ generated by the same rectified flow model is unique. Conversely, the generated $\epsilon$ is unique for a given $x$.


Now, we elucidate how to adapt SDS to the rectified flow to build an initial framework, called  \textit{Vector Field Distillation Sampling} (VFDS). 
The loss, denoted as $\mathcal{L}_{\mathrm{VFDS}}$, can be written as:
\begin{equation}
\nabla_{\theta} \mathcal{L}_{\mathrm{VFDS}}(\phi, x = g(\mathbf{\theta, c})) = \mathbb{E}_{\epsilon, t} 
\left[ 
w(t) \left(v_\phi(x_t, t) - \left(\epsilon - x \right) \right) 
\left( \underbrace{\frac{\partial v_\phi(x_t, t)}{\partial x_t}}_{\text{transformer Jacobian}} 
 \frac{\partial x_t}{\partial x} + 1 \right)
\frac{\partial x}{\partial \theta} \right]
\end{equation}
where $\theta$ is the 3D model parameters, $x = g(\mathbf{\theta, c})$ denotes a rendered image from a camera pose $\mathbf{c}$, $\epsilon$ denotes randomly sampled Gaussian noise, $t \sim U[0,1]$.
Following the convention of the SDS, we omit the transformer Jacobian term for effective training.
Therefore, $\left( \frac{\partial v_\phi(x_t, t)}{\partial x_t} 
 \frac{\partial x_t}{\partial x} + 1 \right)$ becomes a constant and can be absorbed by $w(t)$, so we have
\begin{equation}
\nabla_{\theta} \mathcal{L}_{\mathrm{VFDS}}(\phi, \mathbf{x} = g(\mathbf{\theta, c})) \stackrel{\Delta}{=}
\mathbb{E}_{\epsilon, t} 
\left[ 
w(t)\left(v_\phi(x_t, t) - \left(\epsilon - x \right) \right) 
\frac{\partial x}{\partial \mathbf{\theta}} \right]
\label{eq:Rc-SDS}
\end{equation}

Based on \Eqref{eq:Rc-SDS}, we train a 3D model utilizing a pretrained rectified flow model.
The diffusion model's trajectory is curved~\citep{lipman2022flowmatching}, and the score~\citep{song2020score,dreamfusion} direction varies with different $t$. (see~\Figref{fig:flow_property}(b)).
We denote $\epsilon_\phi(x_t, t)$ as the score function, where $\phi$ represents the parameters of the denoise network, and $x_t$ follows the diffusion forward process. 
In contrast, the trajectory of the rectified flow model~\citep{liu2022rcflow,liu2023instaflow} is straight, and the vector field direction, $v_\phi(x_t, t)$, roughly remains the same for different $t$ under ideal circumstances. 
In our VFDS framework -- where $t \sim U[0,1]$ is used in every optimization step -- the VFDS optimization direction is more consistent, therefore, its convergence speed is faster (please refer to the supplementary material).
However, the over-smoothing issue of SDS still exists.
Therefore, we further analyze the grounding reasons for the over-smoothing issue from the perspective of ODE trajectories.

\begin{wrapfigure}{r}{8cm}
\includegraphics[scale=0.15]{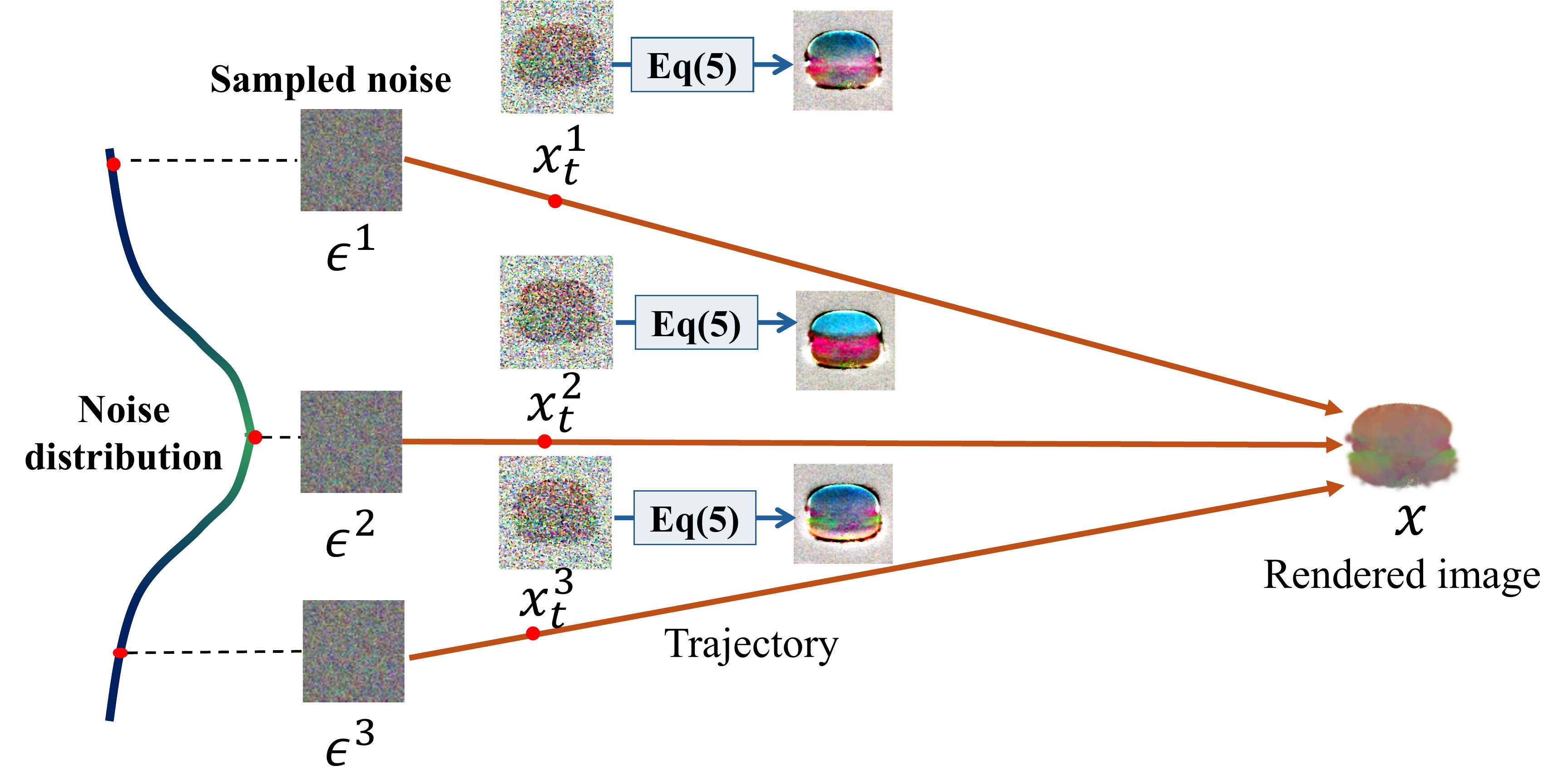}
\vspace{-10pt}
\caption{Illustration for over-smoothing analysis.
An image is coupled with multiple randomly sampled noises, causing the 3D model to learn ODE trajectories.}
\vspace{-7pt}
\label{fig:trajectory}
\end{wrapfigure}

\fakeparagraph{Elucidating SDS Over-smoothing Issue with VFDS}
Because rectified flow is an ODE~\citep{lipman2022flowmatching} model, it has the coupling property.
Now, we analyze the term $\left(v_\phi(x_t, t) - \left(\epsilon - x  \right) \right)$ in \Eqref{eq:Rc-SDS}, where $x_t = t \epsilon + (1-t) x$, $\epsilon \sim \mathcal{N}(\mathbf{0}, \mathbf{I})$ is a random sampled noise and $x$ is the generated image from 3D model.
As shown in \Figref{fig:trajectory}, VFDS randomly samples noise $\epsilon$, which leads to multiple ODE trajectories in the same image $x$.

When camera poses have only mild differences, the rendered images appear nearly identical.
Different ODE trajectories cause inconsistent update directions, which means that directions of $\left(v_\phi(x_t, t) - \left(\epsilon - x  \right) \right)$ are inconsistent. 
\Figref{fig:trajectory} illustrates a toy example.
$\epsilon_1, \epsilon_2, \epsilon_3$ are noise randomly sampled from $\mathcal{N}(\mathbf{0}, \mathbf{I})$ independently. 
According to \Eqref{eq:interpolation}, we can get $x_t^1, x_t^2, x_t^3$, which looks like a blurred image (a hamburger with noise).
Following the above analysis, the same $x$ (rightmost hamburger in \Figref{fig:trajectory}) together with different $\epsilon_1, \epsilon_2, \epsilon_3$ form different trajectories.
During the VFDS training process, the rectified flow model takes $x_t^1, x_t^2, x_t^3$ as input and outputs the estimation of trajectory gradients, as shown in \Figref{fig:trajectory}.
Note that the fitting targets are $\epsilon_1 - x, \epsilon_2 - x, \epsilon_3 - x$ respectively.
The 3D model finally learns the multiple trajectories gradient, causing the over-smoothing issue. 

\subsection{Unique Couple Matching Loss}
\label{sec:ucm_loss}
In Sec.~\ref{sec:vfds-sec}, we have identified the grounding reason for the over-smoothing issue, It is caused by optimizing multiple trajectories during the VFDS training.
On top, we propose a novel Unique Couple Matching (UCM) loss that guides the 3D model to optimize in the same trajectory. 
\textit{Our key idea is to leverage the coupling and reversible properties of rectified flow model}. 

We define the process from $x$ to $\epsilon$ as \textit{push-backward} process, denoted $\#$, which can be written as:
\begin{equation}
\begin{aligned}
    \#_\phi[x] &= x + v_\phi(x,\delta_{T_0}) \Delta_{T_1} + v_\phi(x_{\delta_{T_1}}, \delta_{T_1}) \Delta_{T_2} + \cdots + v_\phi(x_{\delta_{T_{n-1}}}, \delta_{T_{n-1}}) \Delta_{T_n} \\
    x_{\delta_{T_{n-1}}} &= x_{\delta_{T_{n-2}}} + v_\phi(x_{\delta_{T_{n-2}}}, \delta_{T_{n-2}}) \Delta_{T_{n-1}}, n \geq 2
\end{aligned}
\label{eq:push-backward}
\end{equation}

where, $\sum_{i=1}^{n} \Delta_{T_i} = 1$, $\delta_{T_0}=0$. 
And $\#_\phi[x]$ denotes iteratively calculate $v_\phi(x_{\delta_{T_i}},t)$ to backtrack to the $\epsilon$ from $x$ in ~\Eqref{eq:push-backward}.
Due to the reversible property of rectified flow, we can search for a noise $\epsilon$ from $x$ in the VFDS framework.
Additionally, because of the aforementioned coupling property, the search noise is unique.
By replacing randomly sampled noise $\epsilon$ to $\#_\phi[x]$ in \Eqref{eq:Rc-SDS}, our UCM loss is defined as follows:
\begin{equation}
\nabla_{\theta} \mathcal{L}_{\mathrm{UCM}}(\theta, \mathbf{x} = g(\mathbf{\theta, c})) \stackrel{\Delta}{=}
\mathbb{E}_{t} 
\left[ 
w(t)\left(v_\phi(x_t, t) - \left(\#_\phi[x] - x \right) \right) 
\frac{\partial x}{\partial \mathbf{\theta}} \right]
\label{eq:UCM}    
\end{equation}

\begin{equation}
where, x_t = t\#_\phi[x] + (1-t)x
\label{eq:xt}
\end{equation}

As shown in \Figref{fig:pipeline} we can use \textit{push-backward} instead of randomly sampled noise in VFDS, which guides the 3D model to optimize in the same trajectory.

\subsection{FlowDreamer for NeRF and 3D GS}
\label{sec:flowdreamer}
Based on the UCM loss, we propose a novel FlowDreamer
framework. It yields high-fidelity results with richer
textual details and faster convergence. FlowDreamer can be applied to two types of 3D models: 3D Gaussian splatting~\citep{3dgs} and NeRF~\citep{nerf}.

\fakeparagraph{Application to 3D Gaussian Splatting} 
We use points generated by the text-to-3D generator~\citep{nichol2022point_e} as parameter initialization. 
Then, we directly train the 3D GS model using our UCM loss. 
Our FlowDreamer yields high-fidelity results with richer
textual details (see~\Figref{fig:cover}).
It also turns out that our method converges faster than VFDS (please refer to the supplementary material). 

\fakeparagraph{Application to NeRF and the initialization issue} 
For NeRF, however, a direct application does not perform well.
We call this the \textit{initialization issue} of NeRF, as we will explain the reasons below.
When searching the noise $\epsilon$ for a given image $x$, \ie, \textit{push-backward} process, we use the rectified flow model $v_\phi$.
It defines a mapping from data distribution $\pi_0$ to noise distribution $\pi_1$, where $\pi_0$ is the distribution of its pre-training datasets.
The effectiveness of the \textit{push-backward} process depends on that the input distribution for $v_\phi$ should be aligned with or at least approximate to $\pi_0$ or $\pi_1$.
Otherwise, the input lies in an undefined area for $v_\phi$ hence the output is unreasonable.

When training NeRF from scratch, it can hardly generate reasonable images based on its randomly initialized parameters. Therefore, the input distribution (denoted as $\pi^{nf}$) is far from $\pi_0$ and $\pi_1$, causing the rectified flow model $v_\phi$ difficult to estimate the gradient of the ODE trajectory.
To solve this issue, we temporarily use the naive VFDS training to warm up as a remedy.
We view this initialization issue as an open question and advocate further investigations.

As for 3D GS models, the issue does not exist.
The example can be found in the supplementary material, when the prompt ``\textit{A English cottage with stone walls}'' is provided, the NeRF Initialization is simply a gray image, while the output of the 3D GS model (initialized by Point-E~\citep{nichol2022point_e}) resembles a cottage.
This indicates the initial distribution of the 3D Gaussian splatting model is more approximate to $\pi_0$. The result of \textit{push-backward} process is also more effective, as they are closer to the gradient from images output by the trained model.

\vspace{-12pt}
\section{Experiments}

\begin{figure}[t!]
\centering
\includegraphics[width=1.0\linewidth]{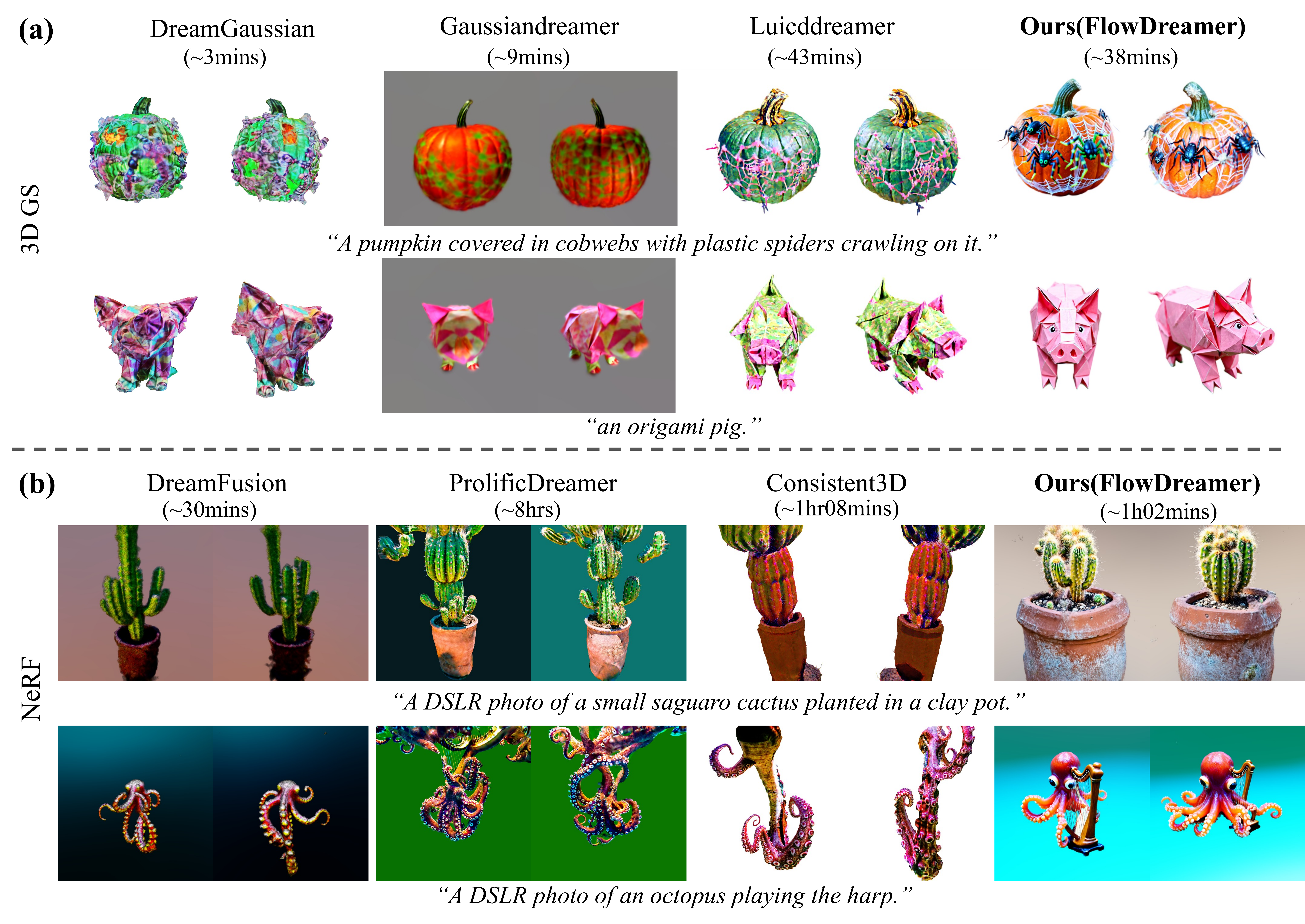} 
\vspace{-15pt}
\caption{Qualitative comparison under 3D GS and NeRF generation setting. Our FlowDreamer generates objects with finer details.}
\vspace{-10pt}
\label{fig:compare}
\end{figure}

\vspace{-5pt}
\subsection{3D Generation Settings}
\vspace{-5pt}
\fakeparagraph{3D Gaussian Splatting Generation}
We compare our FlowDreamer with DreamGaussian~\citep{tang2023dreamgaussian}, GaussianDreamer~\citep{yi2023gaussiandreamer} and LucidDreamer~\citep{liang2023luciddreamer}. 
These 3D GS SoTA baselines are based on their official code by employing the Stable Diffusion 2.1 as the prior.
As shown in \Figref{fig:compare}(a), our method generates objects that match well with the input text prompts and exhibit realistic textures. 
For example, our generated `pumpkin' is of high fidelity, and only our method generates the `spiders', which matches the text prompt 'plastic' for the first prompt. The `origami pig' has rich details, such as its eyes and creases, which are relatively realistic for the second prompt.
Although DreamGaussian~\citep{tang2023dreamgaussian} and GaussianDreamer~\citep{yi2023gaussiandreamer} require comparatively less time, their results are generally subpar.
Our FlowDreamer shows an overall improvement in terms of visual quality and textural details. 

\fakeparagraph{NeRF Generation}
We compare our FlowDreamer with DreamFusion~\citep{dreamfusion}, ProlificDreamer~\citep{wang2024prolificdreamer}, Consistent3D~\citep{wu2024consistent3d} in NeRF.
Other SoTA baselines~\citep{dreamfusion,wang2024prolificdreamer,wu2024consistent3d} reimplemented by Three-studio~\citep{threestudio2023} codebase and employ Stable Diffusion 2.1 for the prior.
As shown in \Figref{fig:compare}(b), our FlowDreamer achieves results with high fidelity and accurate text alignment. 
For example, the `saguaro cactus' and `clay pot' exhibit more detail and greater visual quality for the first prompt. Only FlowDreamer does not render the `octopus' and `harp' as a single object for the second prompt.
Our FlowDreamer takes only more time than DreamFusion~\cite{dreamfusion}, but the quality of DreamFusion's results is limited. (For more results, please refer to the supplementary material.)

\subsection{Quantitative Comparisons}
We use CLIP~\citep{clip} similarity to quantitatively evaluate our method under either NeRF~\citep{nerf} or 3D GS~\citep{3dgs} settings.
\begin{wraptable}{r}{7cm}
\vspace{-10pt}
 \centering
\caption{Quantitative comparisons on CLIP~\citep{clip} similarity with other methods in NeRF generation.}
    \resizebox{\linewidth}{!}{
    \begin{tabular}{l|ccc}
    \hline
    Methods & ViT-B-32 & ViT-L-14 & ViT-g-14 \\
    \hline
    Dreamfusion~\citep{dreamfusion} & 30.13 & 29.70 & 29.49 \\
    Prolificdreamer~\citep{wang2024prolificdreamer} & 32.62 & 32.55 & 31.60 \\
    Consistent3D~\citep{wu2024consistent3d} & 32.34 & 32.56 & 32.01 \\
    \textbf{Ours} & \textbf{34.96} & \textbf{34.19} & \textbf{34.58} \\
    \hline
    \end{tabular}}
\label{tab:nerf_results}
\end{wraptable}
The results of 3D representations with NeRF come from implementation in~\citep{threestudio2023}. The results of 3D representation with 3D GS are from their official implementation. The prompts of NeRF results are from DreamFusion, and the prompts of 3D GS are from LucidDreamer~\cite {liang2023luciddreamer} and ChatGPT.

We randomly choose 26 prompts each to compare in 3D GS and NeRF. We randomly select 12 from the rendered images. The rendered images are from azimuth angles from -180 to 180 degrees with a fixed elevation of 15 degrees for both NeRF and 3D GS. We use three CLIP models from OpenCLIP~\citep{ilharco_gabriel_2021_5143773}, ViT-B-32, ViT-L-14, and ViT-g-14, to calculate the CLIP similarity. Our method demonstrates better CLIP similarities both in NeRF and in 3D GS scenarios. 

\begin{wraptable}{r}{7cm}
\vspace{-20pt}
    \centering
   \caption{Quantitative comparisons on CLIP~\citep{clip} similarity with other methods in 3D Gaussian splatting generation.}
    \resizebox{\linewidth}{!}{
    \begin{tabular}{l|ccc}
    \hline
    Methods & ViT-B-32 & ViT-L-14 & ViT-g-14 \\
    \hline
    DreamGaussian~\citep{tang2023dreamgaussian} & 22.94 & 23.50 & 20.76 \\
    GaussianDreamer~\citep{yi2023gaussiandreamer} & 28.55 & 29.03 & 26.98 \\
    LucidDreamer~\citep{liang2023luciddreamer} & 28.81 & 29.78 & 28.97 \\
    \textbf{Ours} & \textbf{30.70} & \textbf{30.49} & \textbf{30.66} \\
    \hline
    \end{tabular}}
\label{tab:gs_results}
\end{wraptable}
As shown in Tab.~\ref{tab:nerf_results}, our CLIP similarity achieved the best results across all three CLIP models, with the largest margin of 2.57 over the second-best result in ViT-g-14 in NeRF results.
And Tab.~\ref{tab:gs_results} shows that our CLIP similarity also achieved the best results compared to other methods. In particular, it exceeds the LucidDreamer result by 1.89 in ViT-B-32.
\vspace{-5pt}
\subsection{Experimental Insights of our FlowDreamer}
\begin{figure}[t!]
\centering
\includegraphics[width=1.0\linewidth]{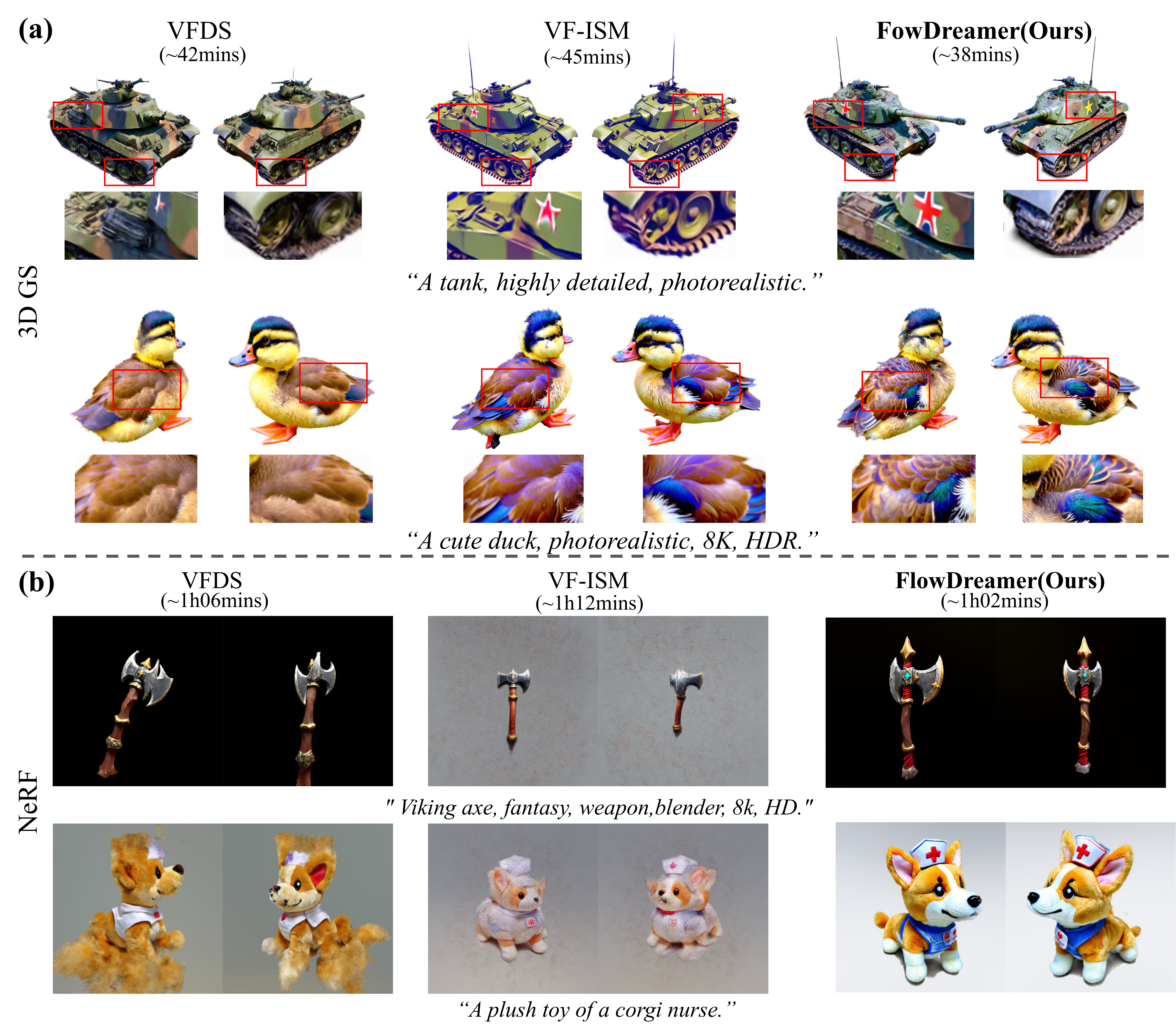} 
\vspace{-20pt}
\caption{Comparison with other baseline methods using the same rectified flow prior under 3D GS and NeRF generation settings. Our 3D results are with richer details and more realistic colors.}
\label{fig:sdv3}
\vspace{-10pt}
\end{figure}

\fakeparagraph{3D generation with rectified flow prior}
To better demonstrate the effectiveness of our method, we replace the SOTA method LucidDreamer's diffusion prior with the rectified flow prior.
\textit{The derivation process is provided in the supplementary material.}
We refer to the ISM loss of LucidDreamer with vector field as VF-ISM.
\Figref{fig:sdv3} demonstrates that our method can generate results with finer details and more realistic textures compared with VFDS and VF-ISM in both 3D GS and NeRF results.
FlowDreamer achieves convergence in 3D GS and NeRF faster than VFDS or VF-ISM, while demonstrating superior details and more realistic shapes.
For instance, \Figref{fig:sdv3}(a) in 3D GS indicate that the tank’s tracks and the duck’s feathers appear heavily blurred in VFDS. Furthermore, the tank's color lacks realism, and the duck's back is somewhat oversaturated in VF-ISM. Our FlowDreamer not only generates the tank and the duck with richer details but also achieves a more realistic overall appearance.
In addition, the \Figref{fig:sdv3}(b) NeRF results reveal that the axe shape and details are improved, whereas the corgi in VFDS exhibits excessive smoothing, and the corgi in VF-ISM presents some noise. The shape and details of our corgi remain relatively satisfactory.

\fakeparagraph{Faster convergence}
Our FlowDreamer can be trained to converge faster than the initial framework VFDS. 
When reaching 1200 steps, our FlowDreamer can generate fine-grained objects whereas VFDS requires more steps to achieve convergence (please refer to the supplementary material). 
When the training step reaches 3000, VFDS still shows much fewer details than FlowDreamer, and it takes less time overall.

\begin{figure}[h]
    \centering
    \includegraphics[width=.95\linewidth]{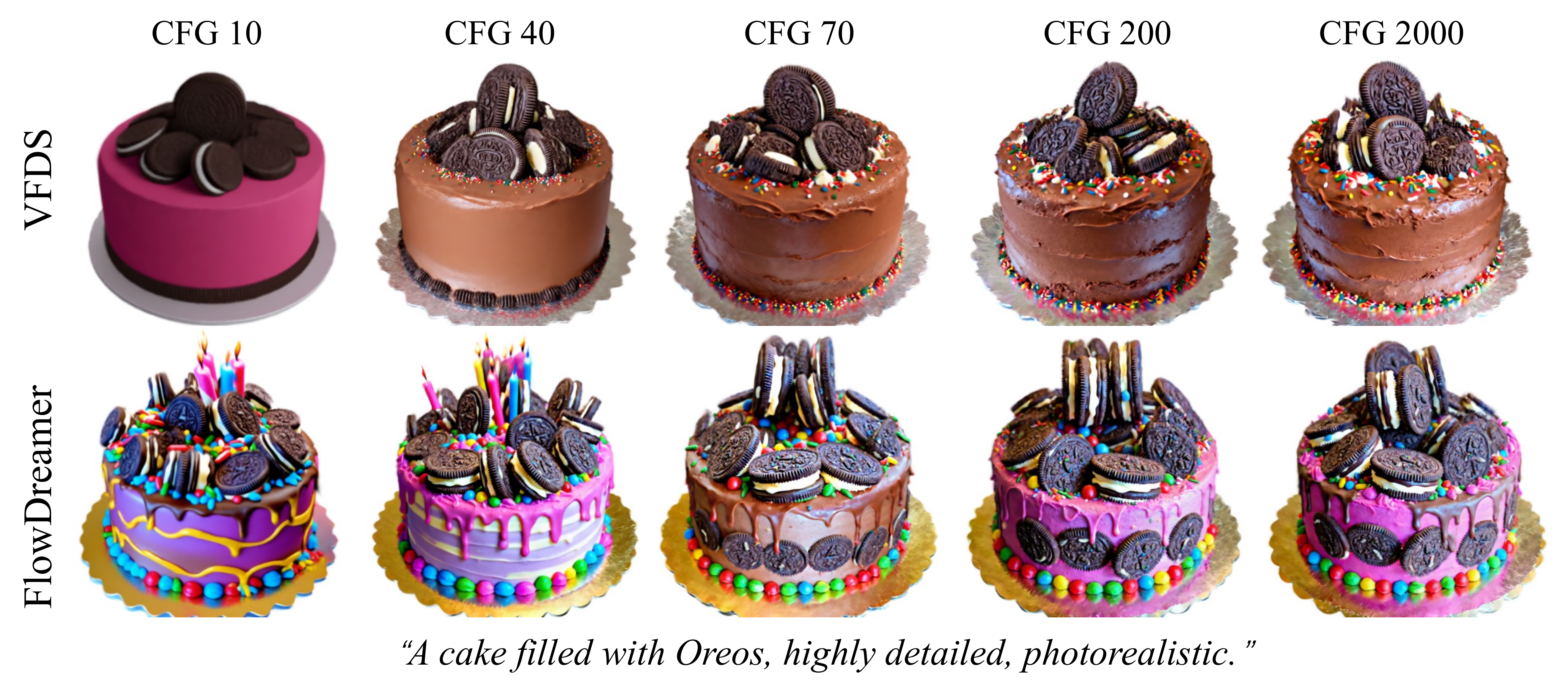} 
     \vspace{-10pt}
    \caption{A comparison between our initial framework VFDS (upper) and FlowDreamer (bottom) with different scales of CFG. 
    The results generated by FlowDreamer contain more detailed features.
    \textbf{Prompt}: \textit{``A cake filled with Oreos, highly detailed, photorealistic.''}}
    \label{fig:cfg_cmp}
\end{figure}

\fakeparagraph{Impacts of different Classifier-Free Guidance (CFG) scales}
We check the impact of CFG (see \Figref{fig:cfg_cmp}).
The results indicate that we achieve good performance across various CFG scales, demonstrating strong robustness to different CFG scales compared with VFDS.
%

%
\begin{figure}[h]
    \centering
    \includegraphics[width=.95\linewidth]{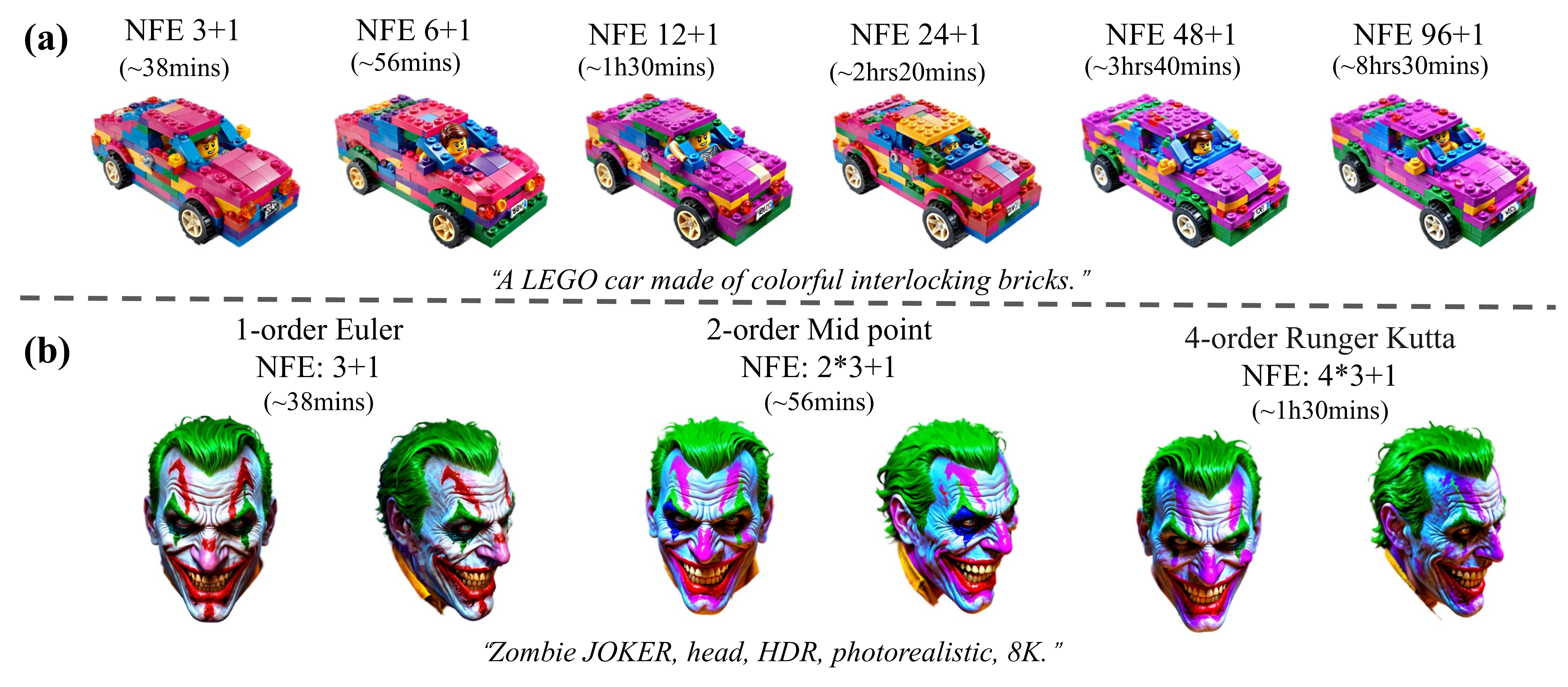} 
    \vspace{-15pt}
    \caption{Impacts of different NFEs and sampling methods. 
    ``NFE N+1'' denotes using $N$ steps of iteration for \textit{push-backward}, and 1 step for gradient calculation. ‘NFE: 2*3+1’ indicates that the second-order method requires 2 inferences per iteration. The total \textit{push-backward} takes 3 iterations.}
    \label{fig:nfe_solver}
    \vspace{-8pt}
\end{figure}

\fakeparagraph{Impacts of different Number of Function Evaluations (NFE)}
As NFE increases, the training time also increases, and the generated objects exhibit more details and more complex structures (see \Figref{fig:nfe_solver}(a)).
For example, the structure complexity of the front hood of the LEGO car gradually increases.
However, Even with a small NFE, wherein the \textit{push-backward} process has a small cost, our method can still train a 3D model with good performance.

\fakeparagraph{Impacts of various sampling methods}
We test three sampling methods, namely first-order Eulder, second-order midpoint, and fourth-order Runge-Kutta.
Experimental results in \Figref{fig:nfe_solver}(b) show that higher-order solvers do not necessarily yield better performance.
For example, the total \textit{push-backward} process takes three iterations, and the Euler method actually produces more realistic results while requiring the least amount of time.
\textit{For more experimental results, please refer to the supplementary material.}
\vspace{-8pt}
\section{Conclusion}
In this paper, we explored a new direction by using the rectified flow model as an alternative prior to text-to-3D generation. 
We developed a mathematical analysis to adapt SDS to rectified flow model, resulting in the initial VFDS framework. However, VFDS still leads to over-smoothing. 
We analyzed this issue from the perspective of ODE trajectories and proposed FlowDreamer, a text-to-3D framework with a new UCM loss. 
Extensive experiments showed that FlowDreamer achieves high-fidelity results with richer details and faster convergence in both NeRF and 3D GS settings. 
We also highlighted open questions, such as initialization issues for NeRF and noise search sampling.
\noindent \textbf{Limitation.}
The Jabus problem still exists; simply adding words like ‘front view,’ ‘back view,’ and ‘side view’ to the prompt is insufficient for supervising the generation view. 
Although we attempt to mitigate this issue using Perpneg~\citep{armandpour2023perpneg}, it still occasionally occurs. We consider solving the Jabus problem thoroughly as a focus for the future work.

\newpage

\clearpage
\bibliography{iclr2024_conference}
\bibliographystyle{iclr2024_conference}

\clearpage
\section*{Appendix}
\appendix
The overview of the Appendix: We provide some illustrations, including the effects of the Transformer Jacobian term, a comparison of the speeds of SDS and VFDS, a comparison of the speeds of VFDS and FlowDreamer, and an example of the initialization difference between NeRF and 3D GS models (see~\ref{sec:illustrations}). Accordingly, we introduce the details of the experiments in~\ref{sec:implementation}. To better compare with other methods, we also conduct a user study to evaluate user preferences in~\ref{sec:user}. The derivation process of VF-ISM with the Rectified Flow prior is discussed in~\ref{sec:vf-ism}. Additionally, we discuss the steps of NFE and sampling methods in~\ref{sec:nfe_sample}. More results include comparisons under 3D GS and NeRF generation settings and additional results from our FlowDreamer (see~\ref{sec:more_results}).

\subsection{Some illustrations}
\label{sec:illustrations}
\begin{figure}[]
\centering
\includegraphics[width=0.9\linewidth]{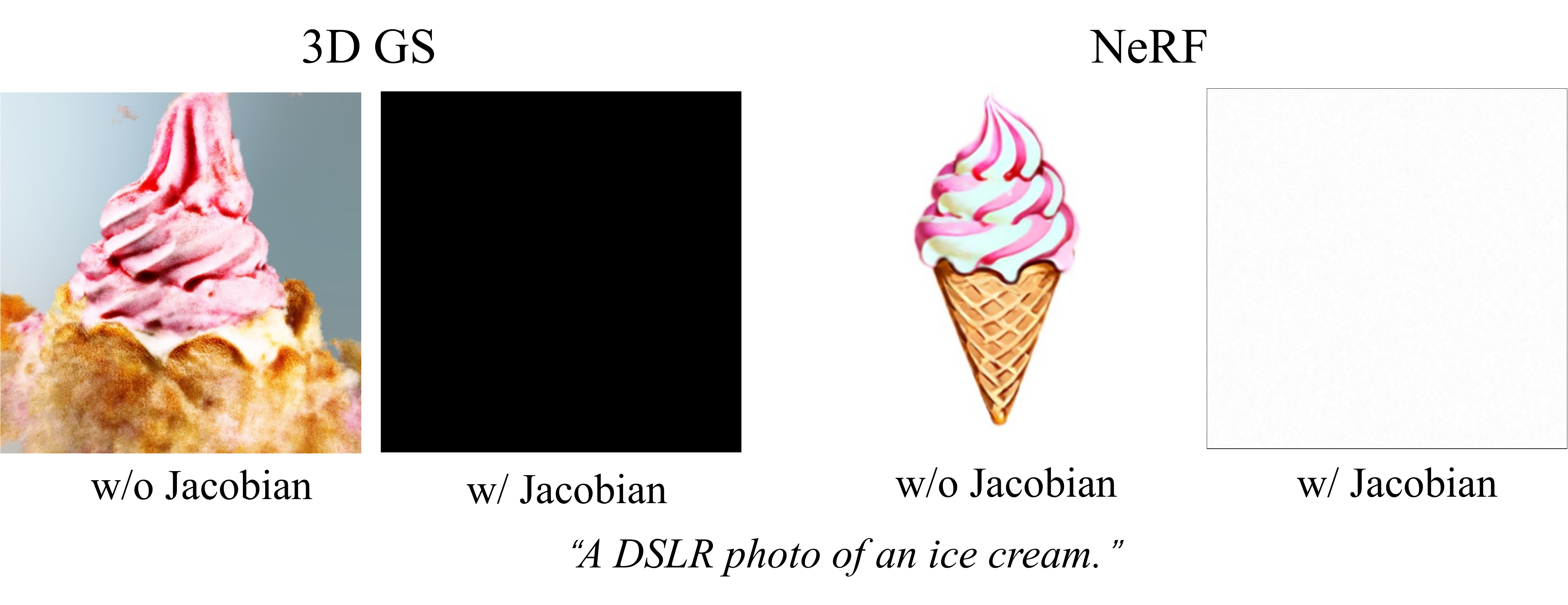} 
     \vspace{-10pt}
\caption{VFDS results of ignoring the Transformer Jacobian. We found that it is difficult to generate meaningful results.}
\label{fig:jacobian}
\end{figure}

\begin{figure}
\centering
\includegraphics[width=1.0\linewidth]{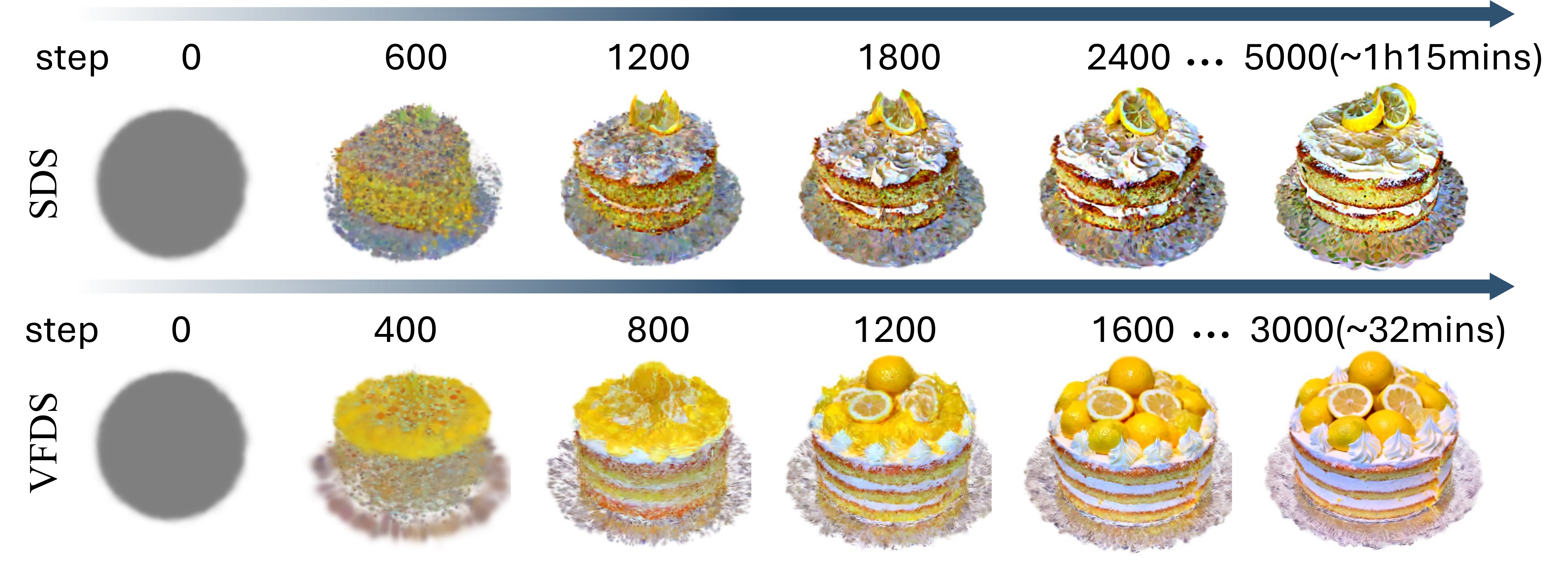}
\caption{A comparison of SDS and VFDS with different steps shows that VFDS converges more quickly and achieves relatively good results in just 3000 steps. \textbf{Prompt:} “A lemon cake.”}
\label{fig:diffusion_vs_flow}
\end{figure}

\begin{figure}[t]
    \centering
    \includegraphics[width=.9\linewidth]{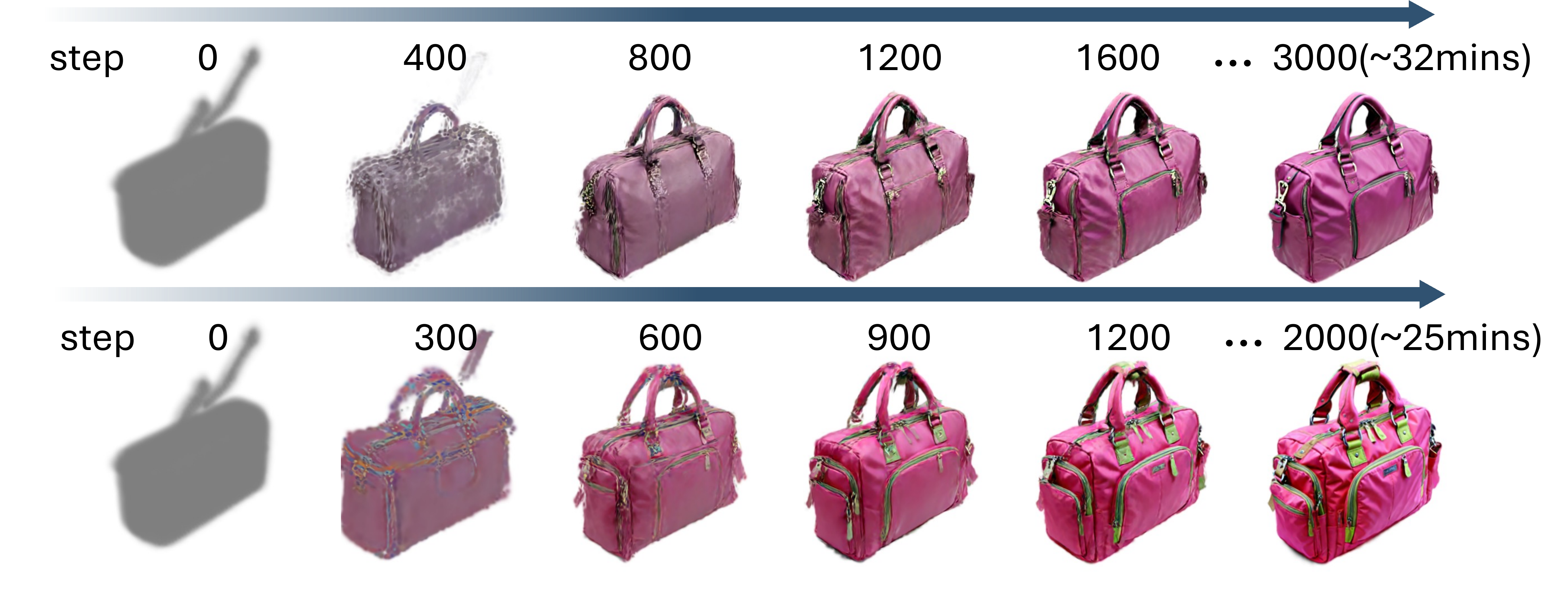} 
     \vspace{-10pt}
    \caption{A comparison of our VFDS and FlowDreamer with different steps. 
    Our method generates images with higher quality than VFDS at the same training steps, indicating a faster convergence speed.
     \textbf{Prompt}: \textit{``A DSLR photo of a large-capacity handbag suitable for travel.''}}
    \label{fig:sds_vs_cp}
\end{figure}

\begin{figure}[t]
\centering
\subfigure[NeRF]{
\includegraphics[width=.45\linewidth]{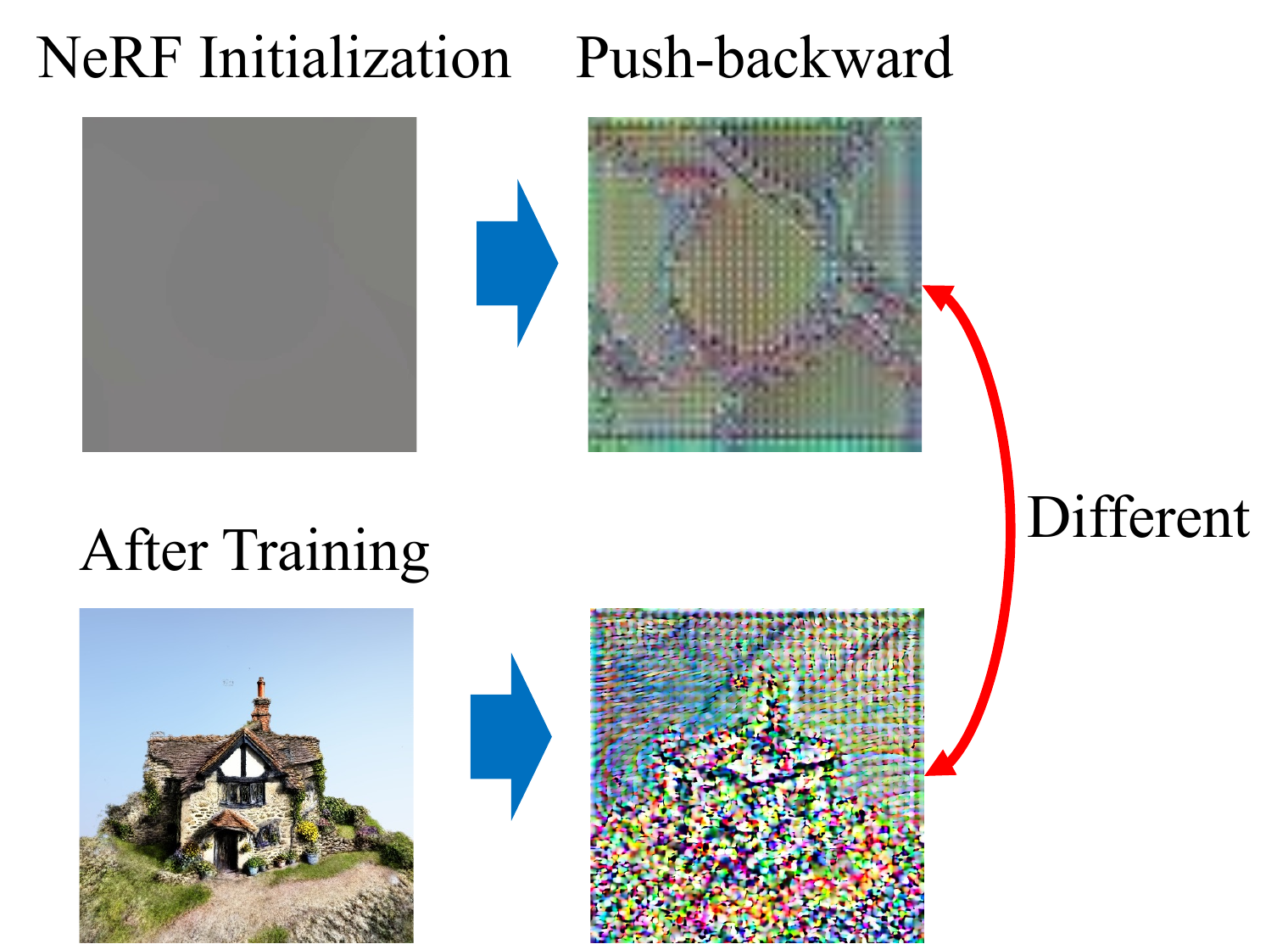} 
}    
\subfigure[3D Gaussian splatting]{
\includegraphics[width=.45\linewidth]{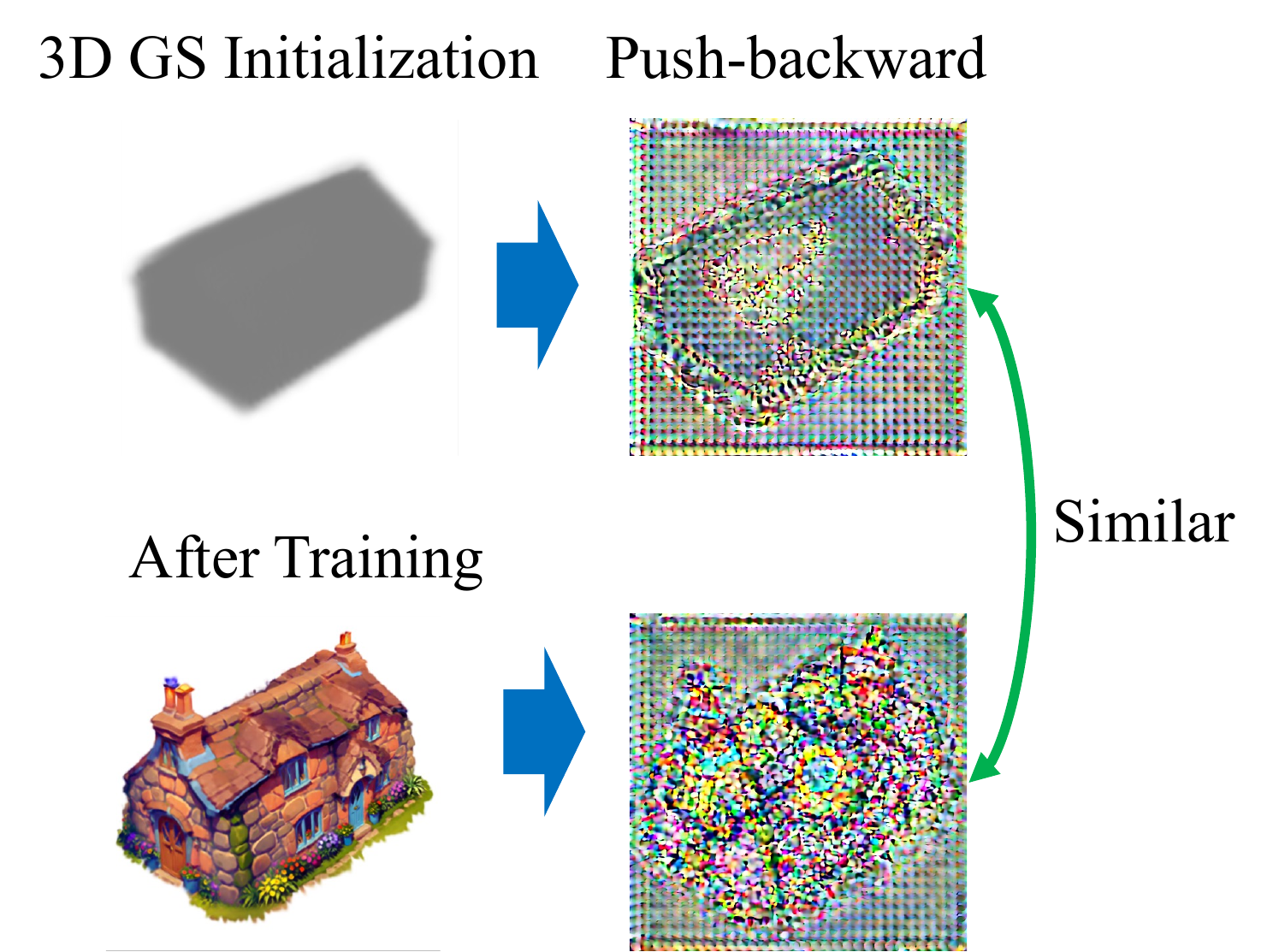} 
}    
\vspace{-12pt}
\caption{Comparison of initialization between NeRF and 3D GS models. 
(\textbf{a}) Images generated by NeRF models before and after training are different, leading to different push-backward results.
(\textbf{b}) As for 3D Gaussian splatting model, the results are quite similar.}
\label{fig:difference_init}
\end{figure}

We provide some illustrations for a better understanding of our paper. 
\Figref{fig:jacobian} illustrates the effects of the Transformer Jacobian term. It is shown that keeping this term leads to training crashes, making it very difficult to generate meaningful 3D objects. Therefore, we omit this term to achieve an effective gradient for optimization. 
\Figref{fig:diffusion_vs_flow} shows a comparison of the convergence speeds of SDS and VFDS. 
\Figref{fig:sds_vs_cp} shows a comparison of the convergence speeds of VFDS and FlowDreamer. 
\Figref{fig:difference_init} illustrates an example of the initialization difference between NeRF and 3D GS models.

\subsection{Implementation Details}
\label{sec:implementation}
We adopt Stable Diffusion 3 (SDv3)~\citep{esser2024sd3} as our Rectified flow model.
To facilitate a better comparison, we will categorize the methods into two types: one where the 3D model is NeRF, and the other where the 3D model is 3D GS.
For NeRF results comparisons, unlike ProlificDreamer~\citep{wang2024prolificdreamer} and Consistent3D~\citep{wu2024consistent3d}, which use multi-stage rendering with normal, geometric, and texture rendering, our method simply uses Instant-NGP~\citep{muller2022instantngp} for rendering. 
It is optimized with a resolution of 256 for the first 5000 steps with a batch size of 1, and then 512 for the subsequent 3000 steps, also with a batch size of 1.
For the warm-up strategy, we use VFDS to optimize over 1200 steps.
The initial framework VFDS need more steps to optimize, which uses 5000 steps for both 256 and 512 resolutions, with the batch size always set to 1.
For 3D Gaussian splatting comparisons, we utilize the pretrained PointE~\citep{nichol2022point_e} to initialize the locations of 3D Gaussians, while other properties of 3D Gaussians adopt random initialization.
Our FlowDreamer uses a batch size of 4 and 3000 iterations, while VFDS uses the same batch size but with 4000 iterations.
The CFG for VFDS is set to 100 on both NeRF and 3D GS, while it is set to 40 for FlowDreamer.

\subsection{User study}
\label{sec:user}

\begin{table}[h]
\centering
\caption{User study of 3D Gaussian splatting methods.}
\resizebox{\linewidth}{!}{
\begin{tabular}{c|cccc}
\hline
Methods & DreamGaussian~\citep{tang2023dreamgaussian} & GaussianDreamer~\citep{yi2023gaussiandreamer} & LucidDreamer~\citep{liang2023luciddreamer} & \textbf{Ours} \\
\hline
Scores & 1.18 & 2.03 & 3.09 & 3.70 \\
\hline
\end{tabular}}
\label{tab:comparison_results_3dgs}
\end{table}

\begin{table}[h]
\centering
\caption{User study of NeRF methods.}
\resizebox{\linewidth}{!}{
\begin{tabular}{c|cccc}
\hline
Methods & DreamFusion~\citep{dreamfusion} & ProlificDreamer~\citep{wang2024prolificdreamer} & Consistent3D~\citep{wu2024consistent3d} & \textbf{Ours} \\
\hline
Scores & 1.46 & 2.70 & 2.18 & 3.65 \\
\hline
\end{tabular}}
\label{tab:comparison_results_nerf}
\end{table}

To compare our method with other comparison methods based on human perception, we conducted a user study involving 36 participants using generated videos from 26 prompts in NeRF and 3D Gaussian splatting, respectively.
Participants viewed four videos simultaneously and assigned scores of 1, 2, 3, and 4 to each video. In each test, the prompts were shown in the title, and participants were instructed to make their decisions based on the degree of alignment of the video with the text, the detail of the video, the color of the video, and the quality of the video. A higher score indicates that users believe the video is better. 
Our FlowDreamer achieves the highest scores among these methods, both in NeRF results and in 3D Gaussian splatting results.

\subsection{VF-ISM Derivation}
\label{sec:vf-ism}
ISM uses DDIM inversion to predict the noisy latent $x_s$ as below.
\begin{equation}
\begin{aligned}
x_{s} &= \sqrt{\bar{\alpha}_{s}} \hat{x}_{}^{s-\delta_T} + \sqrt{1 - \bar{\alpha}_{s}} \epsilon_{\phi}(x_{s- \delta_T}, s-\delta_T, \emptyset)\\
 &= \sqrt{\bar{\alpha}_{s}} \left(\hat{x}_{}^{s - \delta_T} + \gamma(s) \epsilon_{\phi}(x_{s}, s, \emptyset)\right)
\end{aligned}
\end{equation}
where $x_{}^{s-\delta_T} = \frac{1}{\sqrt{\bar{\alpha}_s}} x_{s-\delta_T} - \gamma(s - \delta_T) \epsilon_\phi(x_{s-\delta_T},s-\delta_T, \emptyset)$, 
$\gamma(t) = \frac{\sqrt{1 - \bar{\alpha}_t}}{\sqrt{\bar{\alpha}_t}}$ and $s = t - \delta_T$.
The $\hat{x}^{s-\delta_T}$ is computed using DDIM inference, while $x$ represents the rendered image from the 3D model, as shown below.

\begin{equation}
\begin{aligned}
\hat{x}_{}^{s-\delta_T} &= x - \gamma(\delta_T)\left[\epsilon_\phi(x_{\delta_T},\delta_T, \emptyset) - \epsilon_\phi(x,0,\emptyset) \right] \ldots\\
& \quad - \gamma(s-\delta_T) \left[ \epsilon_\phi(x_{s- \delta_T},s- \delta_T, \emptyset) - \epsilon_\phi(x_{s- 2\delta_T},s- 2\delta_T, \emptyset) \right]
\end{aligned}    
\end{equation}

Next, ISM computes $x_t$ based on $x_s$.
\begin{equation}
\begin{aligned}
x_{t} &= \sqrt{\bar{\alpha}_{t}} \hat{x}_{}^{s} + \sqrt{1 - \bar{\alpha}_{t}} \epsilon_{\phi}(x_s, s, \emptyset)\\
\hat{x}^{s} &= x_{}^{s-\delta_T} - \gamma(s)\left[\epsilon_\phi(x_s,s,\emptyset) - \epsilon_\phi(x_{s-\delta_T},s-\delta_T,\emptyset) \right]
\end{aligned}
\end{equation}

After that, ISM computes the $\tilde{x}^{t}$ with the denoising process.
\begin{equation}
\begin{aligned}
\tilde{x}^{t} = \frac{x_t}{\sqrt{\bar{\alpha}_t}} - &\gamma(t) \epsilon_\phi(x_t,t,y) + \gamma(s) \left[\epsilon_\phi(x_t,t,y) - \epsilon_\phi(x_s,s,y) \right] \\
&+ ... +\gamma(\delta_T) \left[\epsilon_\phi(\tilde{x}_{2\delta_T},2\delta_T,y) - \epsilon_\phi(\tilde{x}_{\delta_T},\delta_T,y)  \right]
\end{aligned}
\end{equation}

ISM computes the $x-\tilde{x}^{t}$ and with the DDIM inversion process.

\begin{equation}
\begin{aligned}
x - \tilde{x}^{t} &= \gamma(t)\left[\epsilon_\phi(\tilde{x}_{t},t,y) - \epsilon_\phi(x_s,s,\emptyset)  \right] + \eta_t \\
\text{where, } \eta_t &= 
+\gamma(s)\left[\epsilon_\phi(\tilde{x}_{s},s,y) 
- \epsilon_\phi(x_{s-\delta_T},s-\delta_T,\emptyset)  \right] 
-\gamma(s)\left[\epsilon_\phi(\tilde{x}_{t},t,y) 
- \epsilon_\phi(x_s,s,\emptyset)  \right] \\ & + ...\\ &+\gamma(\delta_T)\left[\epsilon_\phi(\tilde{x}_{\delta_T},\delta_T,y) 
- \epsilon_\phi(x,0,\emptyset)  \right]
-\gamma(\delta_T)\left[\epsilon_\phi(\tilde{x}_{2\delta_T},2\delta_T,y) 
- \epsilon_\phi(x_{\delta_T},\delta_T,\emptyset)  \right]
\end{aligned}
\end{equation}

To derive VF-ISM, we adapt ISM to rectified flow.
Concretlty, we first conduct \textit{push-backward} operation to compute the $x_s$ with Euler sample method as below.

\begin{equation}
\begin{aligned}
    x_{\delta_T} &= x + v_\phi(x,0,\emptyset)dt \\
    x_{2\delta_T} &= x_{\delta_T} + v_\phi(x_{\delta_T},\delta_T,\emptyset)dt \\
    ...\\
    x_s &= x_{s-\delta_T} + v_\phi(x_{s-\delta_T}, s-\delta_T, \emptyset)dt
\end{aligned}
\end{equation}
where, $dt = \Delta_T$.

Then we use the \textit{push-backward} to compute the $x_t$ with Euler method:
\begin{equation}
    x_t = x_s + v_\phi(x_s, s, \emptyset)ds
\end{equation}
where, $ds = t-s$.

Due to rectified flow, the $\tilde{x}_t^{t}$ can be simplly expressed as below:
\begin{equation}
\begin{aligned}
\tilde{x}^{t} &= x_t - v_\phi(x_t,t,y)ds \\
&- v_\phi(x_s,s,y)dt - v_\phi(x_{s-\delta_T},s-\delta_T,y)dt \\
&- ... - v_\phi(x_{\delta_T},\delta_T,y)dt 
\end{aligned}
\end{equation}

Then we compute the $x-\tilde{x}^{t}$ with \textit{push-backward} with rectified flow prior:
\begin{equation}
\begin{aligned}
x - \tilde{x}^{t} &= \left[ v_\phi(x_t,t,y) - v_\phi(x_s,s,\emptyset) \right]ds + \eta_t \\
\text{where,} \eta_t &= 
\left[ v_\phi(x_s,s,y) - v_\phi(x_{s-\delta_T},s-\delta_T,\emptyset) \right]dt \\
&+... \\
&+\left[ v_\phi(x_{2\delta_T},2\delta_T,y) - v_\phi(x_{\delta_T},\delta_T,\emptyset) \right]dt
\end{aligned}
\end{equation}
Same as ISM, we ignore the $\eta_t$ and use $v_\phi(x_t,t,y) - v_\phi(x_s,s,\emptyset)$ as ISM with Rectified flow prior.


\subsection{Discussion of the steps of NFE and sample methods}
\label{sec:nfe_sample}

\begin{figure}[t!]
\centering
\includegraphics[width=1.1\linewidth]{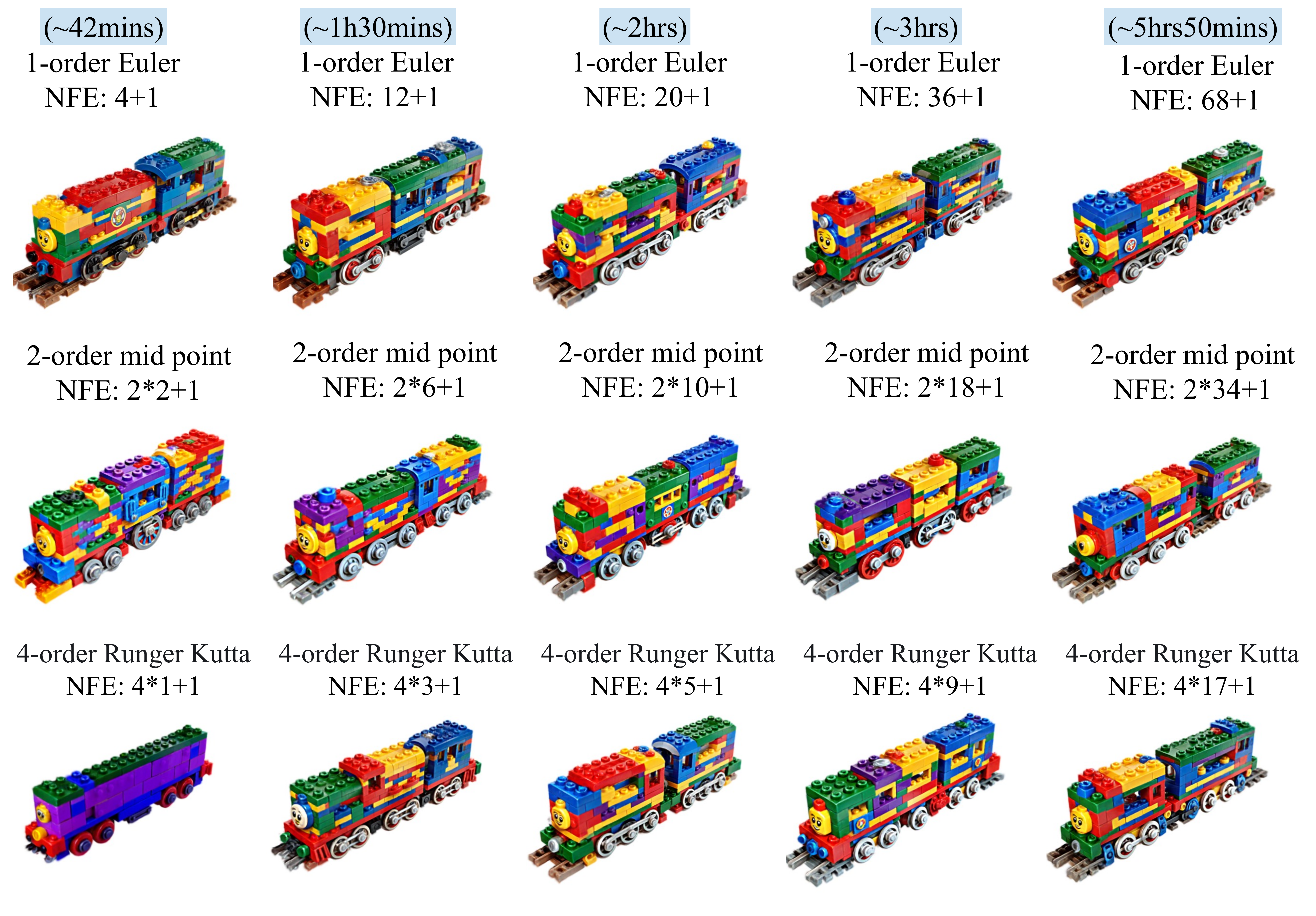} 
\vspace{-15pt}
\caption{The results of different NFEs and sample methods.}
\label{fig:compare_gs}
\end{figure}
As the number of iterations for \textit{push-backward} increases, more training time are required. 
As NFE increases, the LEGO car exhibits more complex structures overall.
When viewed vertically, each column represents results of different sampling methods with the same NFE steps. 
In terms of results for each column, the Euler sampling method demonstrates strong competitiveness, regardless of number of steps.

\subsection{More Results of FlowDreamer}
\label{sec:more_results}

\begin{figure}[]
\centering
\vspace{-2em}
\includegraphics[width=1.0\linewidth]{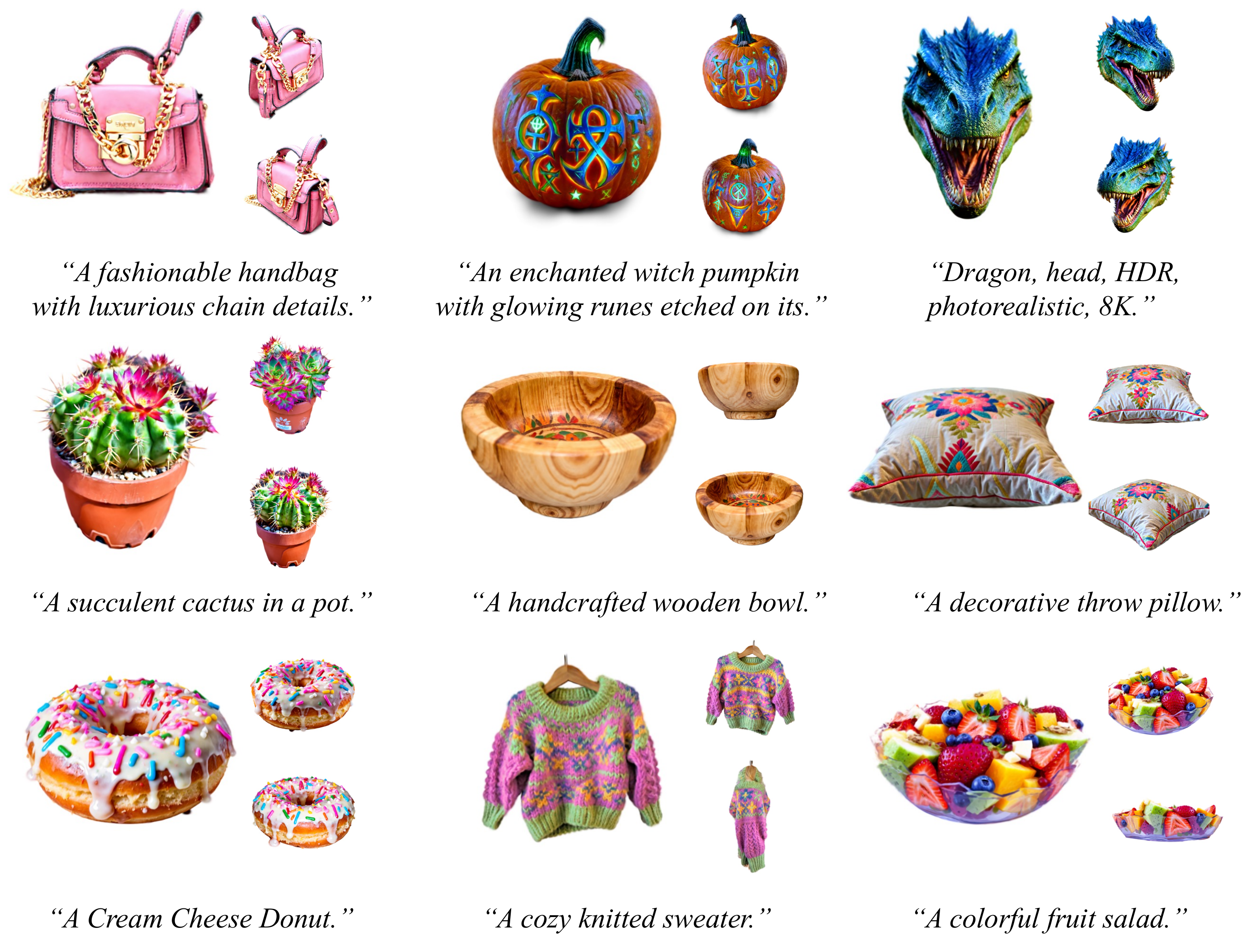} 
\vspace{-2em}
\caption{More results under 3D GS generation setting. Please zoom in for details}
\label{fig:sdv3_3dgs}
\end{figure}

\begin{figure}[b]
\centering
\includegraphics[width=1.0\linewidth]{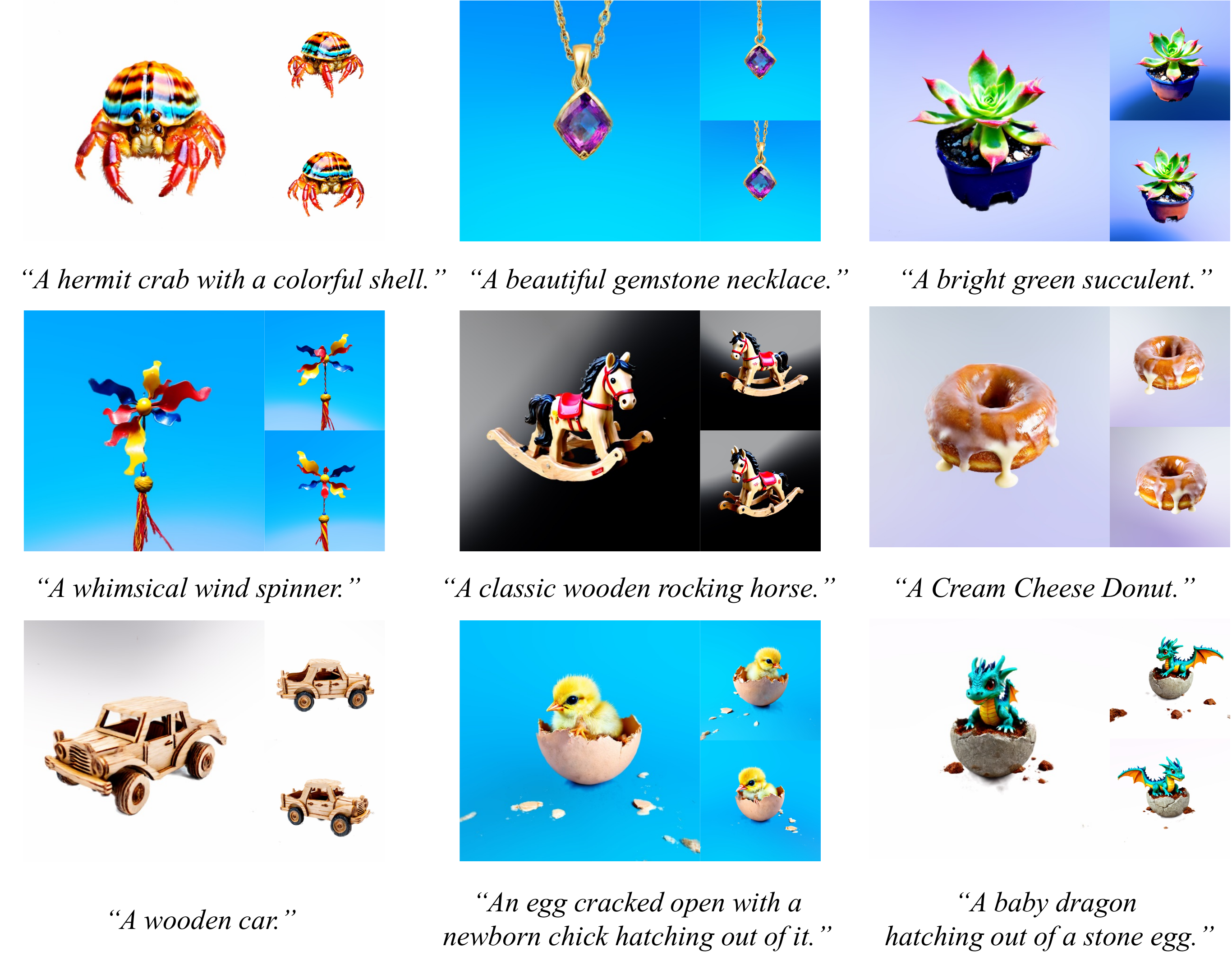} 
\vspace{-15pt}
\caption{More results under NeRF generation setting. Please zoom in for details}
\label{fig:sdv3_nerf}
\end{figure}

As shown in \Figref{fig:sdv3_3dgs} and \Figref{fig:sdv3_nerf}, our FlowDreamer can generate high-fidelity textures and shapes from pretrained rectified flow models both in 3D GS and NeRF.
Our method can produce realistic objects, such as sweaters and wooden bowls, including fantastical ones that are rare in reality, like pumpkins with glowing runes and baby dragons.

\begin{figure}[t!]
\centering
\includegraphics[width=1.1\linewidth]{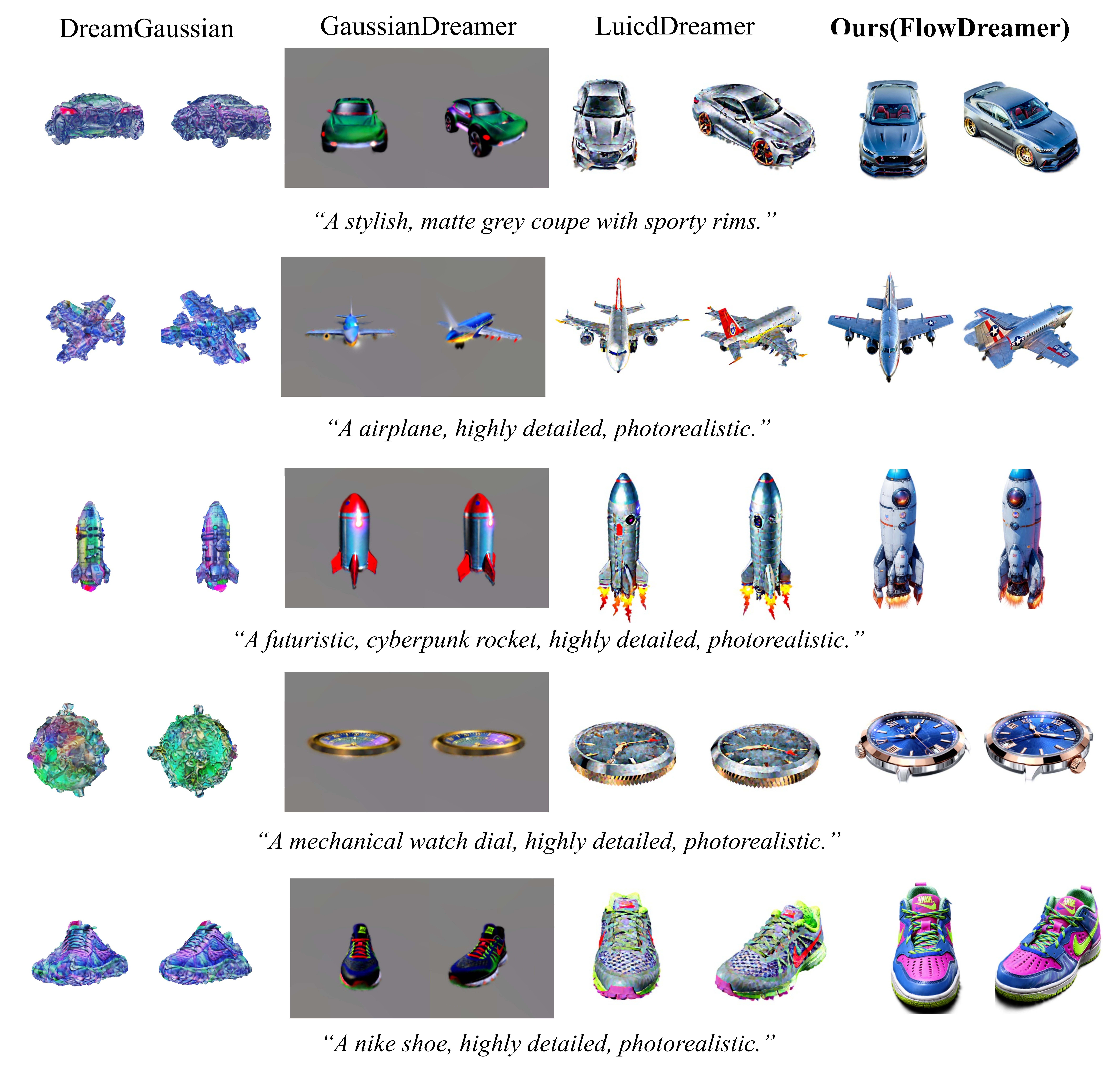} 
\vspace{-15pt}
\caption{More qualitative comparison under 3D GS generation setting.}
\label{fig:sup_compare_3dgs}
\end{figure}

\begin{figure}[t!]
\centering
\includegraphics[width=1.1\linewidth]{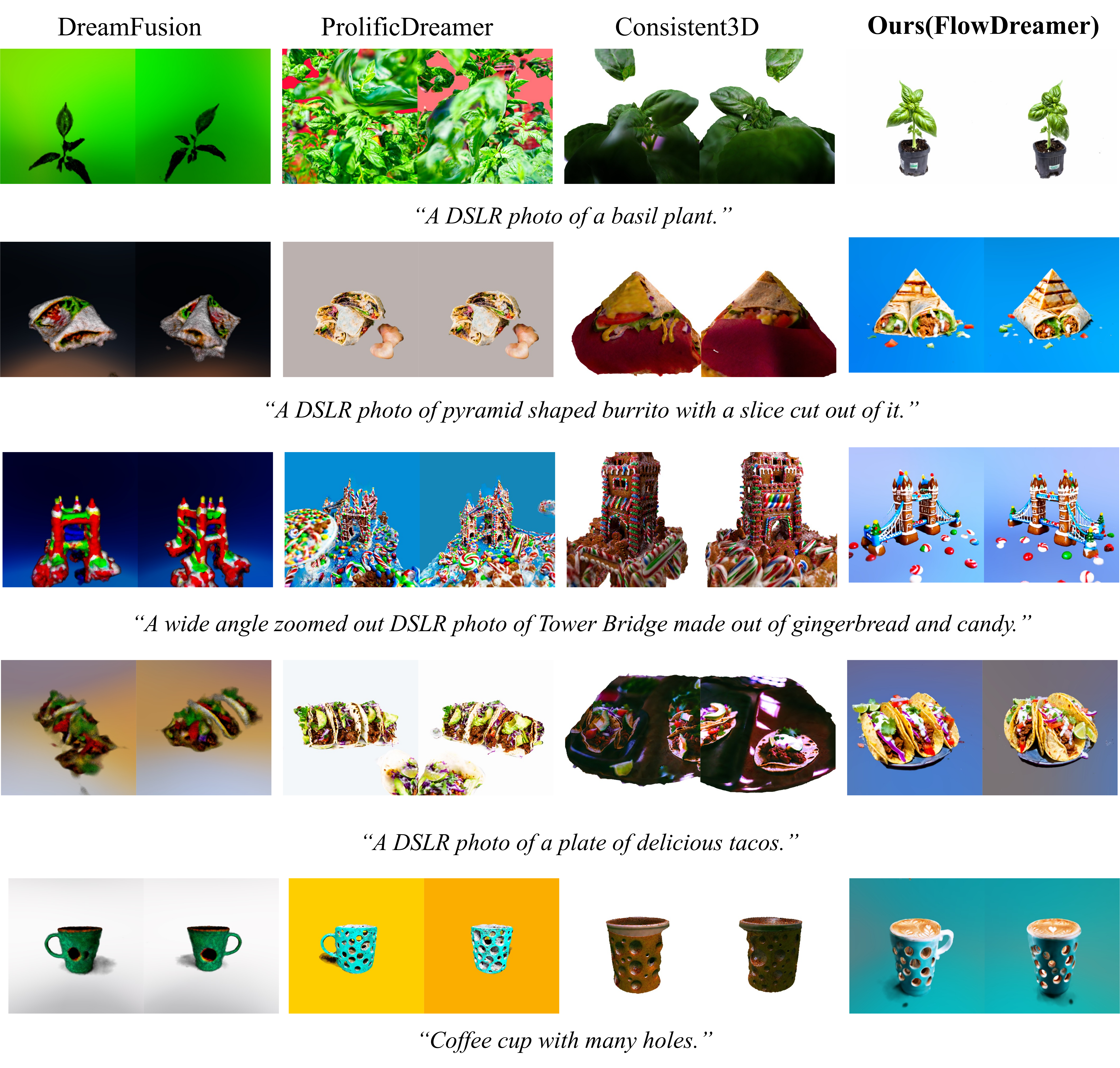} 
\vspace{-15pt}
\caption{More qualitative comparison under NeRF generation setting.}
\label{fig:sup_compare_nerf}
\end{figure}

Comparison with other methods in text-to-3D generation for NeRF and 3D GS, respectively, shows that our FlowDreamer creates 3D objects that match well with the input text prompts, exhibiting high fidelity and intricate details.
The coupe is realistic in both color and shape, and the generated watch dial appears lifelike in 3D GS results (See \Figref{fig:sup_compare_3dgs}).
The burrito's shape and quality are more aligned with the prompts, and the generated tacos are even more realistic in NeRF results (See \Figref{fig:sup_compare_nerf}).

\end{document}